\pdfoutput=1
\documentclass[10pt,twocolumn,letterpaper]{article}

\usepackage{cvpr}              

\usepackage{graphicx}
\usepackage{amsmath}
\usepackage{amssymb}
\usepackage{booktabs}
\usepackage{bm}
\usepackage{algorithm}
\usepackage{algorithmicx}
\usepackage{algpseudocode}
\usepackage[american]{babel}
\usepackage{dsfont}
\usepackage{multirow}

\usepackage[pagebackref,breaklinks,colorlinks]{hyperref}

\newcommand{\squishlist}{
 \begin{list}{$\bullet$}
  { \setlength{\itemsep}{0pt}
     \setlength{\parsep}{1pt}
     \setlength{\topsep}{1pt}
     \setlength{\partopsep}{0pt}
     \setlength{\leftmargin}{1em}
     \setlength{\labelwidth}{1em}
     \setlength{\labelsep}{0.5em} } }

\newcommand{\squishend}{
  \end{list}  }

\def\cvprPaperID{****} 
\def\confName{CVPR}
\def\confYear{2022}

\newcommand{\et}{\textit{et al.}\xspace}

\begin{document}

\title{Salvage of Supervision in Weakly Supervised Object Detection}

\author{Lin Sui$^1$\;\;\;\;\;\;\;\;Chen-Lin Zhang$^{1,2}$\;\;\;\;\;\;\;\;Jianxin Wu$^1$\thanks{J. Wu is the corresponding author. This research was partly supported by the National Natural Science Foundation of China under Grant 61772256 and Grant 61921006.}\\
$^1$State Key Laboratory for Novel Software Technology, Nanjing University, China\\
$^2$4Paradigm Inc., Beijing, China\\
{\tt\small \{suilin0432, zclnjucs, wujx2001\}@gmail.com}
}
\maketitle

\begin{abstract}
  Weakly supervised object detection~(WSOD) has recently attracted much attention. However, the lack of bounding-box supervision makes its accuracy much lower than fully supervised object detection (FSOD), and currently modern FSOD techniques cannot be applied to WSOD. To bridge the performance and technical gaps between WSOD and FSOD, this paper proposes a new framework, Salvage of Supervision (SoS), with the key idea being to harness every potentially useful supervisory signal in WSOD: the weak image-level labels, the pseudo-labels, and the power of semi-supervised object detection. This paper proposes new approaches to utilize these weak and noisy signals effectively, and shows that each type of supervisory signal brings in notable improvements, outperforms existing WSOD methods (which mainly use only the weak labels) by large margins. The proposed SoS-WSOD method also has the ability to freely use modern FSOD techniques. SoS-WSOD achieves 64.4 $m\text{AP}_{50}$ on VOC2007, 61.9 $m\text{AP}_{50}$ on VOC2012 and 16.6 $m\text{AP}_{50:95}$ on MS-COCO, and also has fast inference speed. Ablations and visualization further verify the effectiveness of SoS.
\end{abstract}

\section{Introduction}
\label{sec:intro}

Large-scale datasets with precise annotations are critical in developing detection algorithms, but are expensive to obtain. Thus, weakly supervised object detection~(WSOD), which only needs image-level labels on training images, is popular these days. WSOD has borrowed ideas from fully supervised object detection~(FSOD), such as object proposals~\cite{selectivesearchijcv2013,mcgcvpr2014} and the Fast-RCNN framework~\cite{fastrcnniccv2015}. But modern FSOD has discarded external object proposals and has developed better techniques like Faster-RCNN~\cite{fasterrcnnnips2015} and FPN~\cite{fpncvpr2017}. Furthermore, current WSOD methods mostly use VGG16~\cite{vggiclr2014} as the backbone and Fast-RCNN~\cite{fastrcnniccv2015} as the detector, which confines both accuracy and speed. That is, due to the lack of detailed box-level annotations, WSOD \emph{cannot} enjoy the progress from FSOD. In fact, it has been shown that modern FSOD techniques such as ResNet backbones and RoIAlign will even \emph{deteriorate} WSOD detectors~\cite{enableresneteccv2020}. The weak image-level label is often \emph{the only supervisory signal} utilized for object detection in WSOD, by resorting to a multi-instance recognition setup~\cite{wsddncvpr2016}.

\begin{figure*}
  \centering
  \vspace{-6pt}
  \includegraphics[width=0.75\textwidth]{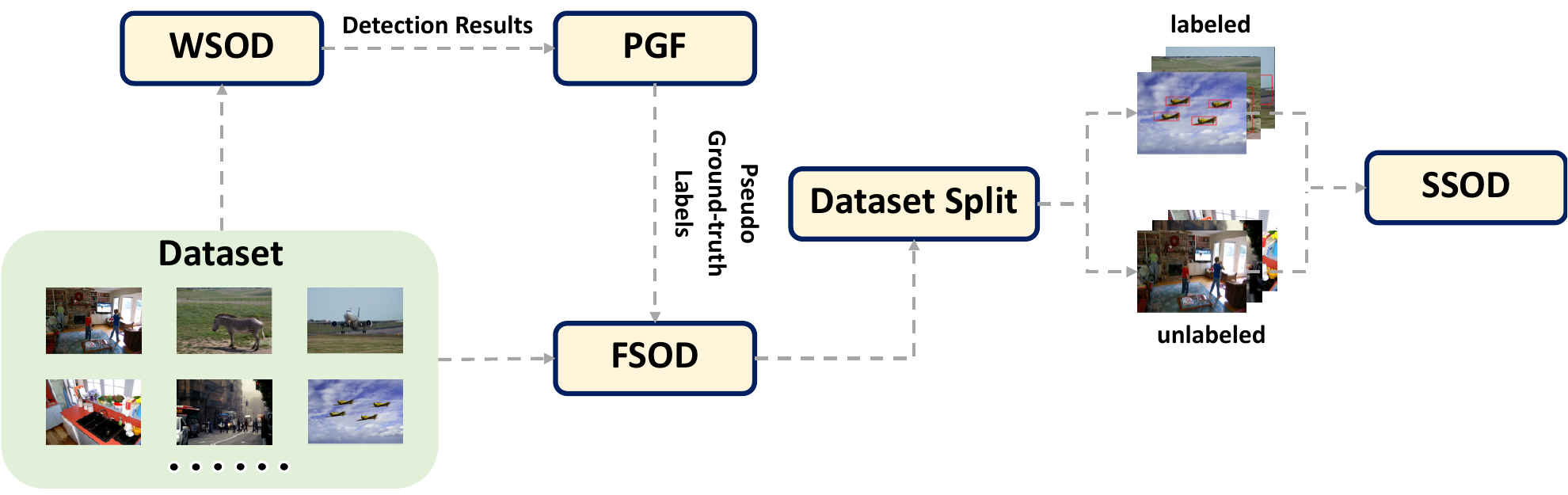}
  \caption{The SoS-WSOD pipeline. Stage 1 trains a weakly supervised detector with only image-level labels. We design PGF to filter its detection results and to generate high-quality pseudo box-level annotations in stage 2, which enables us to train a fully supervised detector. Stage 3 splits the training set into ``clean'' and unlabeled ``noisy'' parts, and trains a detector in a semi-supervised manner.}
  \label{fig-pipeline}
  \vspace{-6pt}
\end{figure*}

In this paper, we argue that WSOD must \emph{fight hard to harness every potential source of supervisory signal}, and should \emph{find a way to utilize the progress in FSOD}. The proposed Salvage of Supervision~(SoS) framework (SoS-WSOD) is illustrated in Fig.~\ref{fig-pipeline}, which has 3 stages. Stage 1 trains a detector with any WSOD method, and we propose an improved OICR~\cite{oicrcvpr2017} as our stage 1. Stage 2 is pseudo-FSOD, where the difficulty is to generate \emph{good} pseudo box-level annotations in order to boost performance and adopt newer FSOD techniques (e.g., ResNet~\cite{resnetcvpr2016}, RoIAlign~\cite{maskrcnn2017}, and FPN~\cite{fpncvpr2017}), i.e., to \emph{salvage} the supervision. This problem has been largely ignored in WSOD, for which we propose a simple but effective solution. Stage 3 is proposed by us, named as SSOD, in which we split the whole dataset into a ``clean'' and a ``noisy'' part, then treat the noisy part as \emph{unlabeled}. That is, we \emph{salvage} additional useful supervisory signals by creating a semi-supervised object detection (SSOD) problem. Hence, we have salvaged supervisory signals out of weak labels by designing novel algorithms to generate high-quality pseudo box-level labels and by creating a semi-supervised learning problem, respectively.

Compared to existing WSOD methods, our SoS-WSOD not only harnesses every potentially useful supervisory signal, but also enables WSOD to fully enjoy both accuracy and speed benefits of modern FSOD methods. Although pseudo FSOD has been tried~\cite{w2fcvpr2018, oicrcvpr2017,slvcvpr2020, cmidniccv2019}, we will show that SoS-WSOD salvages pseudo-supervision of much higher quality. Hence, our contributions are:
\squishlist
    \item We propose SoS-WSOD, a new WSOD framework, showing that we must harness \emph{all} potential supervisory signals in WSOD: generate \emph{high-quality} pseudo-annotations for FSOD, and treat the generated pseudo-label dataset as a noisy dataset to utilize SSOD.
    \item We show that although existing WSOD methods lag far behind FSOD in terms of both accuracy and \emph{technique}, it is very beneficial and feasible to fill this gap. Our pseudo-FSOD enjoys the benefits of all modern FSOD techniques in WSOD, and achieves both higher accuracy and faster speed.
    \item We improve WSOD accuracy by large margins, with 64.4 $m\text{AP}_{50}$ on VOC2007, 61.9 $m\text{AP}_{50}$ on VOC2012, and 16.6 $m\text{AP}_{50:95}$ on MS-COCO. Besides, SoS-WSOD also has fast detection speed.
\squishend

\section{Related Work}
\label{sec:relatedwork}

\textbf{Weakly supervised object detection~(WSOD). } Weakly supervised object detection~(WSOD) seeks to detect the location and type of multiple objects given only image-level labels during training. WSOD methods often utilize object proposals and the multi-instance learning~(MIL) framework. WSDDN~\cite{wsddncvpr2016} was the first to integrate MIL into end-to-end WSOD. OICR~\cite{oicrcvpr2017} proposed pseudo groundtruth mining and an online instance refinement branch. PCL~\cite{pcltpami2018} clustered proposals to improve pseudo groundtruth mining, and C-MIL~\cite{cmilcvpr2019} improved the MIL loss. Recently, MIST~\cite{wetectroncvpr2020} changed the pseudo groundtruth mining rule of OICR, and proposed a Concrete DropBlock module. Zeng \et~\cite{enableresneteccv2020} made the ResNet~\cite{resnetcvpr2016} backbones working in WSOD. CASD~\cite{casdnips2020} proposed self-distillation along with attention to improve WSOD. Some methods~\cite{ocudeccv2020, adjustericcv2021, caticcv2021} proposed to boost WSOD detector performance with the help of fully annotated COCO-60 dataset. 

Some methods tried to adopt modern FSOD techniques into WSOD~\cite{enableresneteccv2020, uwsod2020}. Some methods have used the output of WSOD methods (pseudo box annotations) to retrain WSOD models with FSOD. W2F~\cite{w2fcvpr2018} proposed a pseudo groundtruth excavation and a pseudo groundtruth adaptation module to mine large and complete objects for retraining. However, they directly retrain WSOD models without considering any noise in the generated pseudo-labeled dataset which is bound to be very noisy. In contrast, we propose to reconsider the pseudo-labeled dataset with noisy label training perspective and harness the semi-supervised learning paradigm to squeeze better pseudo-labels.

\textbf{Semi-supervised object detection (SSOD).} SSOD trains a detector with a small set of images with box-level annotations plus many images without any labels. Compared to WSOD, fewer methods have been proposed for SSOD. SSM~\cite{ssmcvpr2018} stitched high-confidence patches from unlabeled  to labeled data. CSD~\cite{csdnips2019} used consistency and background elimination. Recently, STAC~\cite{stacarxiv2020} used strong data augmentation for unlabeled data. Liu \et\cite{unbiasediclr2021} used a teacher-student framework, and ISMT~\cite{ismtcvpr2021} used mean teacher. However, these methods need \emph{an exact split} of labeled and unlabeled data, and \emph{noisy-free box-level annotations} for labeled images, but all these are not available in WSOD. We will generate them from the noisy output of the previous stage in SoS-WSOD.

\textbf{Learning with noisy labels.} As deep neural networks are annotation-hungry, training DNN with noisy labels also attracts much attention, especially in image classification. Some~\cite{jointcvpr2018, pencilcvpr2019} proposed iterative methods to relabel noisy samples by using network predictions. \cite{reweighticml2018, losscorrecticml2018} focused on reweighting. Besides, considering samples with smaller losses as clean ones is also commonly used in many works, such as in~\cite{coteaching2018, dividemixiclr2020}. In SoS-WSOD, we adopt the small loss idea to split the noisy output of stage 2 to split the data into ``clean'' and ``noisy'' parts.

\section{Salvage of Supervision}
\label{sec:soswsod}

\begin{algorithm*}[th]
	\caption{Salvage of Supervision}
	\begin{algorithmic} [1]
		\Statex {\textbf{Input}: Training images $I_{tr}$ and image-level class labels $L_{tr}$, test images $I_{te}$}
		\State \quad Train a WSOD model $W_{wsod}$, and generate pseudo groundtruth bounding boxes ${b}_{tr}$ for training images
		\State \quad Use $I_{tr}$ and ${b}_{tr}$ to train a fully supervised object detector $W_{full}$
		\State \quad Divide $I_{tr}$ into a labeled subset $I_{tr}^*$ with pseudo boxes $b_{tr}^*$ and an unlabeled subset $I_{tr}'$
		\State \quad Use $W_{full}$ to initialize, and learn a semi-supervised $W_{final}$ on $L_{tr}$, $I^*_{tr}$ (with $b^*_{tr}$) and $I'_{tr}$
		\State \textbf{Return:} Use $W_{final}$ to predict the bounding boxes and their class labels for test images
	\end{algorithmic}
	\label{alg:UOD}
	\vspace{-3pt}
\end{algorithm*}

\textbf{Notation.} We first define our notation. A training set $\mathcal{D}_w$ consists of training images $I_{tr}$ and image-level annotations $L_{tr}$. Specifically, each image $\bm{x} \in \mathbb{R}^{h \times w \times 3}$ in $I_{tr}$ has a corresponding label $\bm{y} =[y_1, y_2, \cdots, y_C] \in [0, 1]^C$, where $C$ is the total number of object categories. We will train a detector $W_{final}$ without using any additional annotations.

\textbf{Overview.} Algorithm~\ref{alg:UOD} is the pipeline of the proposed SoS-WSOD. We first train a WSOD detector $W_{wsod}$, which generates pseudo bounding boxes $b_{tr}$. These pseudo supervision signals are used to train an FSOD model $W_{full}$, which can use modern FSOD techniques. Then, we treat the generated pseudo-labeled dataset as a noisy one. With our proposed splitting rule, it is split into an unlabeled subset with $I_{tr}'$ and a ``clean'' labeled subset~(those images with confident pseudo boxes) which consists of $I_{tr}^*$ and $b_{tr}^*$. Finally, we adopt an SSOD method to train the final detector $W_{final}$ on the pseudo labeled dataset.

\subsection{Stage 1: Improved WSOD}

A traditional WSOD detector starts the process. Besides the given image-level annotations $I_{tr}$, most WSOD methods use external object proposals $R$ as extra inputs. Among them, the pipeline of OICR~\cite{oicrcvpr2017} is widely used, which first selects a small number of most confident object proposals $\hat{R}$ as foreground proposals and then refines them by filtering and adding bounding box regression branches.

We propose to improve OICR with two simple changes as our stage 1. First, recent works~\cite{pcltpami2018, wetectroncvpr2020, oimaaai2020, imcfgaaai2021} demonstrate that better proposal mining rules are critical in obtaining higher recall of objects, which are essential for WSOD detectors. For example, MIST~\cite{wetectroncvpr2020} proposed to mine more proposals with low overlap between each other. We find that MIST 
can catch more objects but will also mine a large number of wrong proposals, while OICR is able to mine accurate proposals but ignores many groundtruth instances. Hence, we introduce a mining rule which strikes a balance between recall and precision. In addition, inspired by CASD~\cite{casdnips2020}, we find the multi-input technique is also helpful even \emph{without} using inverted attention and CASD's self-attention transfer. More details are in the appendix.

Our proposed WSOD (stage 1) is a strong baseline (cf. Sec.~\ref{sec:exp}). However, we will also show that SoS-WSOD can achieve excellent performance by adopting a weaker WSOD baseline in stage 1, too.

\subsection{Stage 2: High-quality pseudo boxes for FSOD}

\begin{figure}
  \centering
  \includegraphics[width=0.4\textwidth]{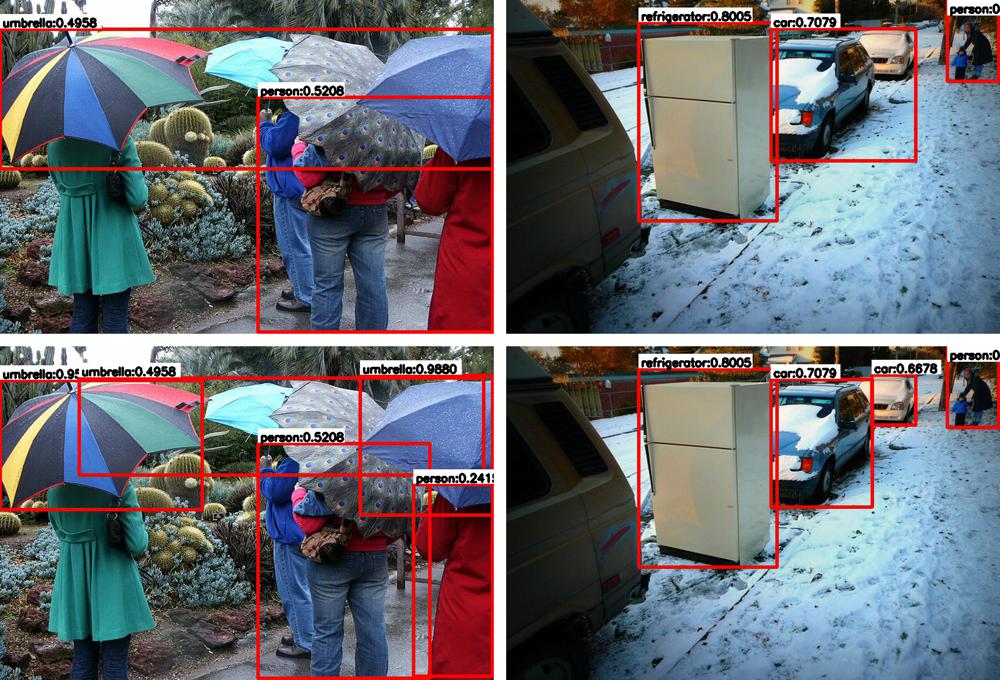}
  \caption{Comparison of W2F~\cite{w2fcvpr2018}~(top) and PGF~(bottom). W2F tends to generate clustered objects in complicated scenarios.}
  \label{fig-w2fvspgf}
  \vspace{-16pt}
\end{figure}

If we are able to output pseudo bounding boxes $b_{tr}$ from stage 1's detector $W_{wsod}$ that are \emph{accurate to some extent}, a subsequent FSOD using these boxes can further improve detection. \cite{oicrcvpr2017} was the first to re-train a WSOD detector by selecting the top-one detection result per class as pseudo groundtruth label, but it will miss a large amount of objects, especially for complicated datasets such as MS-COCO. As will be shown in the ablations in Sec.~\ref{sec:exp}, missed objects will be treated as backgrounds and will even deteriorate the detection accuracy. W2F~\cite{w2fcvpr2018} proposed pseudo groundtruth excavation~(PGE) and pseudo groundtruth adaption~(PGA) to generate pseudo groundtruth from WSOD output. However, W2F only dealt with the VOC datasets, which have a small number of objects per image and the objects are often large in size. Both modules in W2F are designed to mine \emph{large complete} objects, and are not suitable for general detection. Figure~\ref{fig-w2fvspgf} shows that W2F tends to cluster multiple objects into one pseudo-box. Instead, we propose a simple but effective algorithm called pseudo groundtruth filtering~(PGF) to generate high-quality pseudo-boxes from stage 1's WSOD model, whose pipeline is shown in Algorithm~\ref{alg:pgf}.

For each groundtruth class, we only keep the top-scored predictions and those with high confidence~($\ge t_{keep}$, line 6). Then, we remove tiny proposals which are mostly contained inside other proposals in the same category~(lines 8-10). After PGF generates pseudo groundtruth $\hat{P}$, in SoS-WSOD, we are able to use $\hat{P}$ to supervise and train an FSOD detector $W_{full}$ using \emph{modern} FSOD methods (e.g., Faster-RCNN~\cite{fasterrcnnnips2015} + FPN~\cite{fpncvpr2017}). Please note that the impact of our pseudo-FSOD phase is two-fold. First, the retrained WSOD detector gets accuracy \emph{and speed} gains from these salvaged supervisory signals. Besides, now we are able to use \emph{almost all modern FSOD techniques which are previously not applicable in WSOD.} In other words, a WSOD detector now has the \emph{flexibility to select most backbones and architecture as needed in WSOD}, without resorting to extensive efforts (e.g., as in~\cite{enableresneteccv2020}).

We intentionally designed PGF to be very simple in order to achieve both generality and simplicity. In practice, it is also possible to tailor the pseudo groundtruth mining algorithm to the characteristics of the dataset (e.g., as in~\cite{w2fcvpr2018}.)

\begin{algorithm}[t]
    \caption{Pseudo Groundtruth Filtering (PGF)}
    \begin{algorithmic}[1]
        \Statex {\textbf{Input}: boxes $P$ with scores $S$ for an input image (output of stage 1) and its active labels $y_1,\ldots, y_m$, keep threshold $t_{keep}$, containment threshold $t_{con}$}
        \Statex {\textbf{Output}: Pseudo groundtruth boxes $\hat{P}$}
        \State \quad {$\hat{P} = \varnothing$}
        \State \quad \textbf{for} {$i=1,\ \ldots, \ m$}
        \textbf{do}
        \State \quad \quad {$S_i = S[i, :]$ \quad      // get scores for the $i$-th active class}
        \State \quad \quad {$ind_{max},S_i^{max} = \max(S_i)$} \quad  // get index and score of the top proposal
        \State \quad \quad {$P_{i}^{max} = P[ind_{max}, :]$} \quad  // get bounding box for the top proposal
		\State \quad \quad Remove all proposals whose scores $<t_{keep}$, and the remaining boxes form a set $P_i$
		\State \quad \quad $P_{i} = P_{i} \bigcup P_{i}^{max}$
		\State \quad \quad \textbf{for} any two different bounding boxes $u, v$ remaining in $P_i$ \textbf{do}
		\State \quad \quad \quad \textbf{if} $\frac{|u\cap v|}{|v|} \geq t_{con}$ \textbf{then} $P_i = P_i \setminus v$
		\State \quad \quad \textbf{end for}
		\State \quad \quad $\hat{P} = \hat{P} \bigcup P_i$
        \State \quad \textbf{end for}
    \end{algorithmic}
    \label{alg:pgf}
\end{algorithm}

\subsection{Stage 3: Split noisy data for SSOD}

FSOD detectors can bring performance gains to WSOD detectors if a high percentage of the generated pseudo groundtruth are correct. However, noisy or wrong pseudo groundtruth (e.g., missing instances, wrong classification results or inaccurate bounding boxes) are inevitable in WSOD. To deal with this issue, we propose to further salvage supervision by treating the generated pseudo-labeled dataset from the perspective of noisy label learning. After splitting the dataset into ``clean'' labeled part and unlabeled ``noisy'' part, the semi-supervised learning paradigm can be used.

\textbf{Data split.} Many works~\cite{coteaching2018, mentorneticml2018} have shown that noisy annotations will deteriorate the performance. The pseudo groundtruth boxes $\hat{P}$ generated by PGF will inevitably have many noisy ones. As shown in \cite{coteaching2018, noisyanchorfsodcvpr2020, dividemixiclr2020}, a deep network tends to fit clean data first, then gradually memorize noisy ones. Thus, we use the FSOD detector $W'_{full}$~(the detector before performing learning rate decay in the pseudo-FSOD stage) to divide training images $I_{tr}$ into labeled $I_{tr}^*$ (with relatively clean pseudo groundtruth boxes $b_{tr}^*$) and unlabeled ones $I_{tr}'$ (whose pseudo groundtruth boxes are more noisy). In a classification problem, the split is simple~\cite{coteaching2018}: calculate the loss of each training image, and those with smaller loss values are ``clean'' ones. But, in object detection, it is hard to decide whether an image is clean simply based on the sum of all losses of all proposals. We follow the small loss idea but revise it for object detection.

\begin{figure*}
  \centering
  \begin{subfigure}[b]{0.3\linewidth}
    \centering
    \includegraphics[width=1\linewidth]{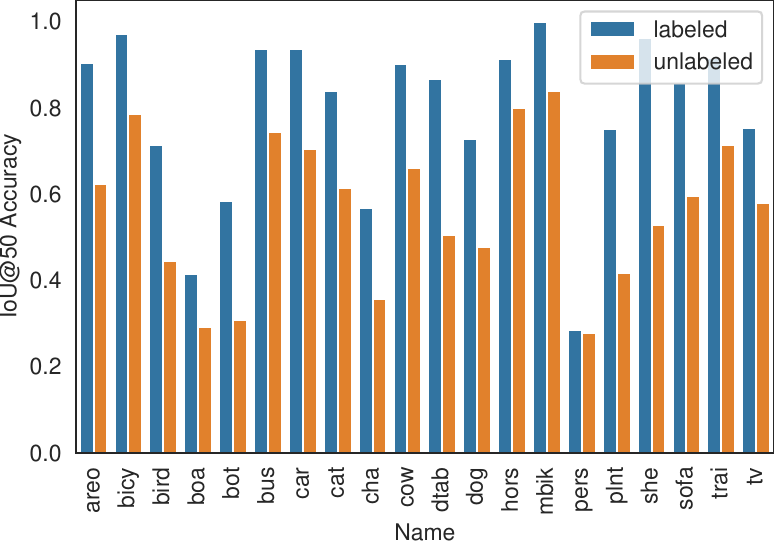}
    \caption{IoU@50}
  \end{subfigure}
  \hspace{24pt}
  \begin{subfigure}[b]{0.3\linewidth}
    \centering
    \includegraphics[width=1\linewidth]{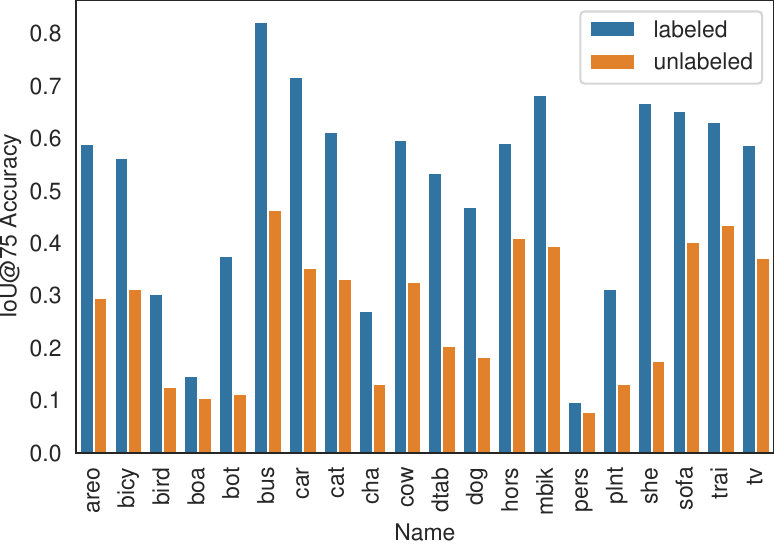}
    \caption{IoU@75}
  \end{subfigure}
  \vspace{-8pt}
  \caption{Per-class accuracy of pseudo groundtruth bounding boxes of the labeled and unlabeled subsets on VOC2007 when $K=2000$.}
  \label{fig-acc}
  \vspace{-10pt}
\end{figure*}

Surely we want to focus on foreground objects, hence we propose the following simple splitting process. In Faster-RCNN, regions of interest (RoIs, denoted as $R$) are divided into foreground and background RoIs according to the IoU between RoIs and pseudo groundtruth boxes. Then, we do \emph{not} calculate losses for background RoIs, and accumulate the RPN losses and RoI losses (both classification and regression branches) of different foreground RoIs. The aggregated loss is the split loss for an input image:
\begin{align}
      \mathcal L_{split}(I) =&\ \frac{1}{N_{pos}}\sum_i \mathds{1}^{R_i}_{fore}\mathcal L_{split}(R_i) \,, \\
      \mathcal L_{split}(R_i) =&\ \mathcal L_{RPN_{Cls}}(R_i) + \mathcal L_{RPN_{Reg}}(R_i) \notag \\
	  &+ \mathcal L_{RoI_{Cls}}(R_i) + \mathcal L_{RoI_{Reg}}(R_i) \,,
      \label{eq: split loss}
\end{align}
where $N_{pos}$ is the number of foreground RoIs, $\mathds{1}^{R_i}_{fore}$ is the indicator function for whether a proposal $R_i$ belongs to foreground RoIs or not, $\mathcal L_{RPN}$ and $\mathcal L_{RoI}$ are RPN and RoI head losses, respectively, and $Cls$ and $Reg$ stand for classification and regression, respectively. We then rank all training images by $\mathcal L_{split}$ and choose images with small loss values as ``clean'' labeled data. For simplicity, we use a hard threshold $K$ to decide sizes of each part. We believe there exist better but more complicated choices such as dynamically using GMM to fit the loss distribution to divide pseudo-labeled dataset. Fig. ~\ref{fig-acc} shows our labeled ``clean'' part are indeed cleaner than the ``noisy'' unlabeled part.

The optimal choice of $K$ varies depends on the size and difficulty of datasets, and we provide a rule-of-thumb for it. We find that traditional WSOD performs well for easy images which will be split as ``clean''. Thus, we set $K$ around the number of images which has a single class label (or few for datasets with many object categories). Ablation studies in Sec.~4 verify the effectiveness of such a rule-of-thumb.

\textbf{Semi-supervised detection.} Now we can perform semi-supervised detection. Unbiased Teacher~\cite{unbiasediclr2021} is an effective semi-supervised detector, whose key idea is a teacher-student pair updated by mutual learning. It first trains a detector using only labeled data (i.e., burn-in) and then uses it to initialize both the student and the teacher detectors. In the mutual learning phase, the teacher will dynamically generate pseudo labels for unlabeled data with weak data augmentation. The student will learn from both well-annotated labeled data and strong augmented unlabeled data with the generated pseudo labels. The teacher will receive updates from the student via exponential moving average.

But, clean data is \emph{not} available in WSOD. We use the Unbiased Teacher pipeline with a few changes and improvements. First, the pseudo-FSOD training is actually a burn-in process, and we do not need an additional burn-in stage. Then, the student learns by minimizing
\begin{equation}
    \mathcal{L} = \mathcal{L}_{sup} + \lambda_u \mathcal{L}_{unsup} \,,
    \label{eq: SSOD learn}
\end{equation}
where the student will learn from both labeled ($\mathcal{L}_{sup}$) and unlabeled ($\mathcal{L}_{unsup}$) data, and $\lambda_u$ is the weight to balance the unsupervised and supervised loss terms. Specifically, $\mathcal{L}_{sup}$ and $\mathcal{L}_{unsup}$ are defined as follows:
\begin{align}
        \mathcal{L}_{sup} =& \sum_i \mathcal{L}_{RPN_{Cls}}(\bm{x}_i^s, \bm{b}_i^s) + \mathcal{L}_{RPN_{Reg}}(\bm{x}_i^s, \bm{b}_i^s) \notag\\ 
		    &+ \mathcal{L}_{RoI_{Cls}}(\bm{x}_i^s, \bm{b}_i^s) + \mathcal{L}_{RoI_{Reg}}(\bm{x}_i^s, \bm{b}_i^s)\,, \\
        \mathcal{L}_{unsup} =& \sum_{i} \mathcal{L}_{RPN_{Cls}}(\bm{x}_i^u, \bm{b}_i') + \mathcal{L}_{RoI_{Cls}}(\bm{x}_i^u, \bm{b}_i')\,,
    \label{eq: loss details}
\end{align}
where $\bm{x}^s_i$ and $\bm{b}_i^s$ are images and pseudo groundtruth boxes in the ``clean'' subset $I_{tr}^*$. $\bm{x}_i^u$ and $\bm{b}_i'$ are images in the unlabeled subset $I_{tr}'$ and pseudo groundtruth dynamically generated by the teacher. $\mathcal{L}_{sup}$ is for labeled data only. For $\mathcal{L}_{unsup}$, the teacher generates pseudo labels for the student with weak data augmentations, then the student uses strong data augmentations along with pseudo labels to calculate it. We believe the predictions of the teacher are less accurate than annotations for the ``clean'' data, so $\mathcal{L}_{unsup}$ only contains the classification loss. In other words, all the regression branches are only learned with ``clean'' labeled data. 

Pseudo boxes generated by the teacher for unlabeled images are not always accurate. Hence, different from regular SSOD, we utilize the image-level labels (i.e., another salvage of supervision) by filtering out false positive pseudo labels, which brings additional benefits to WSOD. Finally, the student detector updates its weights according to Eq.~\ref{eq: SSOD learn}, and the teacher receives its update from the student through exponential moving average~(EMA).

\section{Experiments}
\label{sec:exp}

We evaluated our SoS-WSOD on three standard WSOD benchmark datasets: VOC2007~\cite{vocijcv2010}, VOC2012~\cite{vocijcv2010} and MS-COCO~\cite{cocoeccv2014}. VOC2007 has 2501 training, 2510 validation and 4952 test images. VOC2012 contains 5717 training, 5823 validation, and 10991 test images. MS-COCO contains around 110,000 training and 5000 validation images. Following the common WSOD evaluation protocol, we use training and evaluation images to train our model on VOC2007 and VOC2012, and evaluate on the test images. For MS-COCO, we train our model on the training images and evaluate on the validation images. We use $m\text{AP}_{50:95}, m\text{AP}_{50}$ and $m\text{AP}_{75}$ as evaluation metrics for both MS-COCO and VOC2007. For VOC2012, since labels for test images are not released, we report $m\text{AP}_{50}$ results returned by the official evaluation server.

\subsection{Implementation details}

We use PyTorch on RTX3090 GPUs, and our code will be released soon. Backbone models are pretrained on ImageNet. It is worth noting that WSOD methods lag behind FSOD in terms of backbone and other techniques. For example, state-of-the-art WSOD methods still use VGG16 as the backbone, while FSOD methods use better architectures. Extra efforts are needed in order to adapt modern backbones to WSOD~\cite{enableresneteccv2020}. Instead, in stage 2 and 3 our SoS-WSOD has the freedom to choose backbones and detection architectures. For simplicity and efficiency, we use FPN with ResNet50 backbone as the FSOD detector in our main experiments \emph{without any extra handling}. We also use VGG16 as backbone and remove FPN for fair comparisons.

Details of our improved OICR for the WSOD training stage (stage 1) are available in the appendix. In PGF (Algorithm~\ref{alg:pgf}), we set $t_{keep} = 0.2$ and $t_{con}=0.85$ for \emph{all} datasets. Although generating pseudo groundtruth labels with TTA (Test Time Augmentation) leads to higher accuracy, the high computational cost (1.5/3/33 hours on VOC2007/2012/MS-COCO) makes it hard to use in large-scale datasets. To keep the same setting in all experiments, we do \emph{not} use TTA in Algorithm~\ref{alg:pgf}.

In both stages 2 and 3, we keep \emph{all} hyperparameters except $K$ the same as the \emph{default} hyperparameters in~\cite{detectron22019} and~\cite{unbiasediclr2021}, respectively. We reduce the total training iterations to reduce the training cost. As for $K$ in the data splitting process, we use our rule-of-thumb to set it as 2000 and 30000 for VOC2007 and MS-COCO, respectively. More details can be found in the appendix.

\subsection{Comparison with state-of-the-art methods}

We compare our method with state-of-the-art WSOD methods, with the results reported in Tables~\ref{tab:sota-voc} and~\ref{tab:sota-coco}. All results are reported with TTA. Our improved WSOD baseline (stage 1 of SoS-WSOD) reaches $55.0\%$ $m\text{AP}_{50}$, $52.5\%$ $m\text{AP}_{50}$ and $11.9\%$ $m\text{AP}_{50:95}$ on VOC2007, VOC2012 and MS-COCO, respectively, which is already pretty strong.

\begin{table}
    \centering
    \resizebox{0.45\textwidth}{!}{
    \begin{tabular}{cccc}
        \hline
        \multicolumn{1}{c|}{\multirow{2}{*}{Method}}     & \multicolumn{1}{c|}{\multirow{2}{*}{Backbone}} & \multicolumn{1}{c|}{VOC07}             & VOC12             \\
        \multicolumn{1}{c|}{}                            & \multicolumn{1}{c|}{}                          & \multicolumn{1}{c|}{$m\text{AP}_{50}$} & $m\text{AP}_{50}$ \\ \hline 
        \multicolumn{4}{c}{Pure WSOD}                                                                                                                                  \\ \hline
        \multicolumn{1}{c|}{PCL~\cite{pcltpami2018}}                         & \multicolumn{1}{c|}{VGG16}                     & \multicolumn{1}{c|}{43.5}              & 40.6              \\ \hline
        \multicolumn{1}{c|}{W2F~\cite{w2fcvpr2018}}                         & \multicolumn{1}{c|}{VGG16}                     & \multicolumn{1}{c|}{52.4}              & 47.8              \\ \hline
        \multicolumn{1}{c|}{Pred Net~\cite{prednetcvpr2019}}                    & \multicolumn{1}{c|}{VGG16}                     & \multicolumn{1}{c|}{52.9}              & 48.4              \\ \hline
        \multicolumn{1}{c|}{C-MIDM + FR~\cite{cmidniccv2019}}                 & \multicolumn{1}{c|}{VGG16}                     & \multicolumn{1}{c|}{53.6}              & 50.3              \\ \hline
        \multicolumn{1}{c|}{SLV + FR~\cite{slvcvpr2020}}                    & \multicolumn{1}{c|}{VGG16}                     & \multicolumn{1}{c|}{53.9}              & -                 \\ \hline
        \multicolumn{1}{c|}{WSOD2~\cite{wsod2iccv2019}}                       & \multicolumn{1}{c|}{VGG16}                     & \multicolumn{1}{c|}{53.6}              & 47.2              \\ \hline
        \multicolumn{1}{c|}{IM-CFB~\cite{imcfgaaai2021}}                      & \multicolumn{1}{c|}{VGG16}                     & \multicolumn{1}{c|}{54.3}              & 49.4              \\ \hline
        \multicolumn{1}{c|}{MIST~\cite{wetectroncvpr2020}}                        & \multicolumn{1}{c|}{VGG16}                     & \multicolumn{1}{c|}{54.9}              & 52.1              \\ \hline
        \multicolumn{1}{c|}{CASD~\cite{casdnips2020}}                        & \multicolumn{1}{c|}{VGG16}                     & \multicolumn{1}{c|}{56.8}              & 53.6              \\ \hline
        \multicolumn{1}{c|}{SoS-WSOD~(stage 1)}     & \multicolumn{1}{c|}{VGG16}                     & \multicolumn{1}{c|}{55.0}              & 52.5              \\ \hline
        \multicolumn{1}{c|}{SoS-WSOD~(stage 1+2+3)} & \multicolumn{1}{c|}{VGG16}                     & \multicolumn{1}{c|}{\textbf{60.3}}              & \textbf{57.7}              \\ \hline
        \multicolumn{1}{c|}{SoS-WSOD~(stage 1+2+3)} & \multicolumn{1}{c|}{ResNet50}                  & \multicolumn{1}{c|}{\textbf{64.4}}              & \textbf{61.9}              \\ \hline
        \multicolumn{4}{c}{WSOD with transfer~(using fully annotated COCO-60)}                                                                                        \\ \hline
        \multicolumn{1}{c|}{OCUD + FR~\cite{ocudeccv2020}}                   & \multicolumn{1}{c|}{ResNet50}                  & \multicolumn{1}{c|}{60.2}              & -                 \\ \hline
        \multicolumn{1}{c|}{LBBA~\cite{adjustericcv2021}}                        & \multicolumn{1}{c|}{VGG16}                     & \multicolumn{1}{c|}{56.6}              & 55.4              \\ \hline
        \multicolumn{1}{c|}{CaT~\cite{caticcv2021}}                         & \multicolumn{1}{c|}{VGG16}                     & \multicolumn{1}{c|}{59.2}              & -                 \\ \hline
    \end{tabular}
    }
    \caption{Comparison on PASCAL VOC.}
    \label{tab:sota-voc}
\end{table}

\begin{table}
    \centering
    \resizebox{0.45\textwidth}{!}{
    \begin{tabular}{c|c|ccc}
        \hline
        \multirow{2}{*}{Method} & \multirow{2}{*}{Backbone} & \multicolumn{3}{c}{MS-COCO}                                  \\
                                &                           & $m\text{AP}_{50:95}$ & $m\text{AP}_{50}$ & $m\text{AP}_{75}$ \\ \hline
        PCL~\cite{pcltpami2018}                     & VGG16                     & 8.5                  & 19.4              & -                 \\ \hline
        C-MIDN~\cite{cmidniccv2019}                  & VGG16                     & 9.6                  & 21.4              & -                 \\ \hline
        WSOD2~\cite{wsod2iccv2019}                   & VGG16                     & 10.8                 & 22.7              & -                 \\ \hline
        MIST~\cite{wetectroncvpr2020}                    & VGG16                     & 12.4                 & 25.8              & 10.5              \\ \hline
        CASD~\cite{casdnips2020}                    & VGG16                     & 12.8                 & 26.4              & -                 \\ \hline
        SoS-WSOD~(stage 1)       & VGG16                     & 11.9                 & 24.2              & 10.7              \\ \hline
        SoS-WSOD~(stage 1+2+3)   & VGG16                     & \textbf{15.5}                 & \textbf{30.5}              & \textbf{14.3}              \\ \hline
        SoS-WSOD~(stage 1+2+3)   & ResNet50                  & \textbf{16.6}                 & \textbf{32.8}              & \textbf{15.2}              \\ \hline
    \end{tabular}}
    \caption{Comparison on MS-COCO.}
    \label{tab:sota-coco}
    \vspace{-12pt}
\end{table}

For a fair comparison, we used VGG16 as backbone and did not use modern FPN architecture in stages 2 and 3. By harnessing all possible supervision signals, SoS-WSOD finally reaches $60.3\%$ and $57.7\%$ $m\text{AP}_{50}$ on VOC2007 and VOC2012, which outperforms previous methods by large margins. On MS-COCO, SoS-WSOD reaches $15.5\%$ $m\text{AP}_{50:95}$, $30.5\%$ $m\text{AP}_{50}$ and $14.3\%$ $m\text{AP}_{75}$, which outperforms previous methods significantly, too.

When further adopting modern techniques in FSOD, with the help of ResNet50 backbone and FPN architecture, SoS-WSOD further reaches $64.4\%$ and $61.9\%$ $m\text{AP}_{50}$ on VOC2007 and VOC2012, respectively. On MS-COCO, accuracy is boosted to $16.6\%$ $m\text{AP}_{50:95}$, $32.8\%$ $m\text{AP}_{50}$ and $15.2\%$ $m\text{AP}_{75}$.

Recently, some methods~\cite{ocudeccv2020, adjustericcv2021, caticcv2021} leverage the well-annotated MS-COCO-60 dataset~(removing the 20 categories in VOC). As shown in Table~\ref{tab:sota-voc}, they have higher accuracy than pure WSOD methods because of the additional cross-domain data. However, SoS-WSOD achieves higher accuracy than them without resorting to these additional data.

\subsection{Ablation studies and visualization}

\textbf{Are salvaged supervision signals useful?} Tables~\ref{tab:sota-voc} and \ref{tab:sota-coco} already show that both pseudo boxes (stage 2) and semi-supervised detection (stage 3) notably improve detection accuracy on all 3 datasets. Furthermore, Tables~\ref{tab:ablation-VOC07} to \ref{tab:ablation-COCO} show results on VOC2007, VOC2012 and MS-COCO, respectively. Our improved WSOD (stage 1 of SoS-WSOD) reaches $54.1\%$, $51.8\%$ and $23.6\%$ $m\text{AP}_{50}$ on VOC2007, VOC2012 and MS-COCO, respectively. After pseudo-FSOD (stage 2), $m\text{AP}_{50}$ is improved by $3.5\%$, $2.1\%$ and $3.9\%$, respectively. Finally, another $5.1\%$, $5.7\%$ and $3.1\%$ higher $m\text{AP}_{50}$ are boosted by stage 3, respectively. For the stricter $m\text{AP}_{50:95}$ metric on MS-COCO, stage 2 and 3 bring $18.1\%$ and $15.5\%$ relative improvements, respectively.

\begin{table}
    \centering
	\footnotesize
	\begin{tabular}{c|c|c|c|c|c}
        \hline
        WSOD & PGF & SSOD & $m\text{AP}_{50:95}$ & $m\text{AP}_{50}$ & $m\text{AP}_{75}$ \\ 
        \hline
        \checkmark &  &  & 26.2 & 54.1 & 22.8 \\ \hline
        \checkmark & \checkmark &      & 27.3 & 57.6 & 22.5  \\ \hline
        \checkmark & \checkmark & \checkmark & \textbf{31.6} & \textbf{62.7} & \textbf{28.1}  \\ \hline
    \end{tabular}
    \caption{Ablations of SoS-WSOD stages on VOC2007.}
    \label{tab:ablation-VOC07}
    \vspace{-4pt}
\end{table}

\begin{table}
    \centering
	\footnotesize
	\begin{tabular}{c|c|c|c}
        \hline
        WSOD & PGF & SSOD & $m\text{AP}_{50}$ \\ 
        \hline
        \checkmark &  &  & 51.8 \\ \hline
        \checkmark & \checkmark &      & 53.9 \\ \hline
        \checkmark & \checkmark & \checkmark & \textbf{59.6} \\ \hline
    \end{tabular}
    \caption{Ablations of SoS-WSOD stages on VOC2012.}
    \label{tab:ablation-VOC12}
\end{table}

\begin{table*}[t]
    \centering
	\footnotesize
    \begin{tabular}{c|c|c|c|c|c|c|c|c}
    \hline
    WSOD & PGF & SSOD & $m\text{AP}_{50:95}$ & $m\text{AP}_{50}$ & $m\text{AP}_{75}$ & $m\text{AP}_{S}$ & $m\text{AP}_{M}$ & $m\text{AP}_{L}$ \\ 
    \hline
    \checkmark &     &      & 11.6 & 23.6 & 10.4 & 2.3 & 11.9 & 20.2 \\ \hline
    \checkmark & \checkmark &      & 13.7 & 27.5 & 12.2 & 3.8 & 15.1 & 22.0 \\ \hline
    \checkmark & \checkmark & \checkmark & \textbf{15.5} & \textbf{30.6} & \textbf{14.4} & \textbf{5.4} & \textbf{16.8} & \textbf{24.6} \\ \hline
    \end{tabular}
    \caption{Ablations of SoS-WSOD stages on MS-COCO.}
    \label{tab:ablation-COCO}
\end{table*}

\textbf{Compatibility with other WSOD methods.} SoS-WSOD is compatible with various WSOD methods in stage 1 and can also improve a weaker WSOD method. We tested the original OICR method in stage 1, and results with TTA are in Table ~\ref{tab:dif-wsod}. The basic OICR model gets 50.2 $m\text{AP}_{50}$ on VOC2007.  With SoS-WSOD, such a model finally reaches 59.9 $m\text{AP}_{50}$. These results demonstrate the flexibility and effectiveness of our SoS-WSOD.

\begin{table}
    \centering
	\footnotesize
    \begin{tabular}{c|ccc}
        \hline
        Method          & $m\text{AP}_{50:95}$ & $m\text{AP}_{50}$ & $m\text{AP}_{75}$ \\ \hline
        baseline~(OICR)  & 24.1                 & 50.2              & 19.5              \\ \hline
        baseline~(OICR+) & 27.1                 & 55.0              & 24.8              \\ \hline
        SoS-WSOD~(OICR)  & 28.5                 & 59.9              & 23.8          \\ \hline
        SoS-WSOD~(OICR+)     & 32.6                 & 64.4              & 29.6              \\ \hline
    \end{tabular}
    \caption{Ablations of adopting different WSOD model in stage 1. OICR+ is our improved OICR.}
    \label{tab:dif-wsod}
\end{table}

\textbf{Details about accuracy gains in stage 3.} As stated in~\cite{stacarxiv2020,unbiasediclr2021}, a mixture of weak and strong data augmentation and EMA updating are essential for SSOD methods. As shown in Table~\ref{tab:details}, in addition to gains from strong augmentation and EMA updating, our method always brings significant gains. As for models use EMA and strong data augmentation but do not adopt the splitting rule~(i.e., the third stage), we use the same~(mix weak and strong) data augmentation used in the student branch of stage 3 and maintain another model by EMA in stage 2. SoS-WSOD works well for VGG16 w/o FPN, even when strong augmentation and EMA updating have minor improvements.

\begin{table}
    \centering
    \resizebox{0.47\textwidth}{!}{
    \begin{tabular}{c|c|c|c|ccc}
        \hline
        Backbone & FPN       & EMA \& Aug       & Split Rule & $m\text{AP}_{50:95}$ & $m\text{AP}_{50}$ & $m\text{AP}_{75}$ \\ \hline
        VGG16    & $\times$  & $\times$   & $\times$   & 26.8                 & 56.2              & 22.3              \\ \hline
        VGG16    & $\times$  & $\checkmark$ & $\times$   & 27.3                 & 56.6              & 22.8              \\ \hline
        VGG16    & $\times$  & $\checkmark$ & $\checkmark$  & \textbf{28.8}                 & \textbf{59.0}              & \textbf{24.3}              \\ \hline
        ResNet50    & $\checkmark$ & $\times$ & $\times$   & 27.3                 & 57.6              & 22.5              \\ \hline
        ResNet50    & $\checkmark$ & $\checkmark$ & $\times$   & 29.6                 & 59.6              & 25.7              \\ \hline
        ResNet50    & $\checkmark$ & $\checkmark$ & $\checkmark$  & \textbf{31.6}                 & \textbf{62.7}              & \textbf{28.1}              \\ \hline
    \end{tabular}
    }
    \caption{Detailed accuracy gains in stage 3. The columns mean whether FPN, strong data augmentation, EMA, and the proposed data splitting rule are used or not.}
    \label{tab:details}
	\vspace{-8pt}
\end{table}

\textbf{Effectiveness of PGF.} As shown in Table \ref{tab:sota-voc}, some WSOD methods tried to add a re-train stage. Most of them follow \cite{oicrcvpr2017} to re-train a WSOD detector by selecting the top-one detection result per class as pseudo groundtruth labels. However, the performance of the retrained detector model starts to saturate with the increasing performance of the WSOD model, and pseudo-label generation method is one of the most important reasons. Experimental results in Table~\ref{tab:pgf-compare} show the effectiveness of the proposed PGF. Besides, the widely adopted method~(Top-One in Table~\ref{tab:pgf-compare}) failed to get benefits from modern techniques. When training with RPN and FPN, missed objects will be treated as backgrounds, which will deteriorate localization. Besides, conspicuous objects also have more chance to become the top score proposal which would cause imbalanced anchor allocation and inadequate training in FPN. As for W2F~\cite{w2fcvpr2018}, we have slightly better results on VOC07, and are far superior than it on the more complicated MS-COCO, as W2F is designed for large complete objects in VOC.

\begin{table}
    \centering
    \resizebox{0.45\textwidth}{!}{
    \begin{tabular}{c|c|ccc}
        \hline
        Method    & Dataset & $m\text{AP}_{50:95}$ & $m\text{AP}_{50}$ & $m\text{AP}_{75}$ \\ \hline
        Top-One~\cite{oicrcvpr2017}   & VOC07   & 25.8                 & 53.4              & 22.2              \\ \hline
        W2F$^*$~\cite{w2fcvpr2018}       & VOC07   & 27.2                 & 57.0              & 22.4              \\ \hline
        PGF~(Ours) & VOC07   & \textbf{27.3}                 & \textbf{57.6}              & \textbf{22.5}              \\ \hline
        W2F$^*$       & MS-COCO & 12.6                 & 24.1              & 11.7              \\ \hline
        PGF~(Ours) & MS-COCO & \textbf{13.7}                 & \textbf{27.5}              & \textbf{12.2}             \\ \hline
    \end{tabular}
     }
    \caption{Results of different pseudo groundtruth mining algorithms. $^*$ means that we implemented W2F because its code is not available.}
    \label{tab:pgf-compare}
	\vspace{-12pt}
\end{table}

\textbf{Splitting rule vs. random splitting.} To demonstrate the effectiveness of the proposed splitting rule in stage 3, we compare it with the random splitting strategy. As shown in Table~\ref{tab:split-rule}, adopting random splitting is notably worse than our proposed method. However, random splitting can still surpass simply adopting EMA update and strong data augmentations in stage 2 by a clear margin, which demonstrates the importance of salvaging useful supervisory signals.

\begin{table}
    \centering
    \resizebox{0.45\textwidth}{!}{
    \begin{tabular}{c|ccc}
        \hline
        Method                      & $m\text{AP}_{50:95}$ & $m\text{AP}_{50}$ & $m\text{AP}_{75}$ \\ \hline
        Stage 2 w/ EMA \& Aug & 29.6                 & 59.6              & 25.7              \\ \hline
        Stage 3 w/ random splitting  & 30.2                 & 61.1              & 26.3              \\ \hline
        Stage 3 w/ ours              & \textbf{31.6}                 & \textbf{62.7}              & \textbf{28.1}              \\ \hline
    \end{tabular}
    }
    \caption{Comparison of our splitting rule and random splitting}
    \label{tab:split-rule}
\end{table}

\textbf{Longer training schedule for WSOD.} Counting all 3 stages in, SoS-WSOD does require more training iterations. Hence, we double the training iteration of the WSOD stage for a further fair comparison. However, we find that $m\text{AP}_{50}$ will drop from $54.1\%$ to $52.7\%$ due to overfitting.

\textbf{Size of the labeled subset in SSOD.} In the SSOD stage (stage 3), we split a dataset into labeled and unlabeled subsets. The number of pseudo labeled images, $K$, is a hyperparameter. When we treat a small number of images as ``clean'' labeled ones, severe class imbalance will deteriorate the performance. However, when splitting most images as labeled, the performance will collapse using fully pseudo annotated labels. As shown in Table~\ref{tab:ablation-k}, $K=2000$ is a suitable choice for VOC2007. These results also demonstrate the effectiveness of the rule-of-thumb we proposed. We use $K=2000$ in all our experiments on VOC2007, and double the size to 4000 on VOC2012. Following the proposed rule-of-thumb, for MS-COCO, we use $K=30000$.

\begin{table}
    \centering
	\footnotesize
    \begin{tabular}{c|c|c|c}
    \hline
    K    & $m\text{AP}_{50:95}$ & $m\text{AP}_{50}$ & $m\text{AP}_{75}$\\ 
    \hline
    1000 & 31.2 & \textbf{63.2} & 26.8 \\ \hline
    2000 & \textbf{31.6} & 62.7 & \textbf{28.1}\\ \hline
    3000 & 31.0 & 62.3 & 27.2\\ \hline
    \end{tabular}
    \caption{Effects of $K$ in stage 3 on VOC2007.}
    \label{tab:ablation-k}
    \vspace{-8pt}
\end{table}

\begin{figure*}
	\centering
	\includegraphics[width=0.86\textwidth]{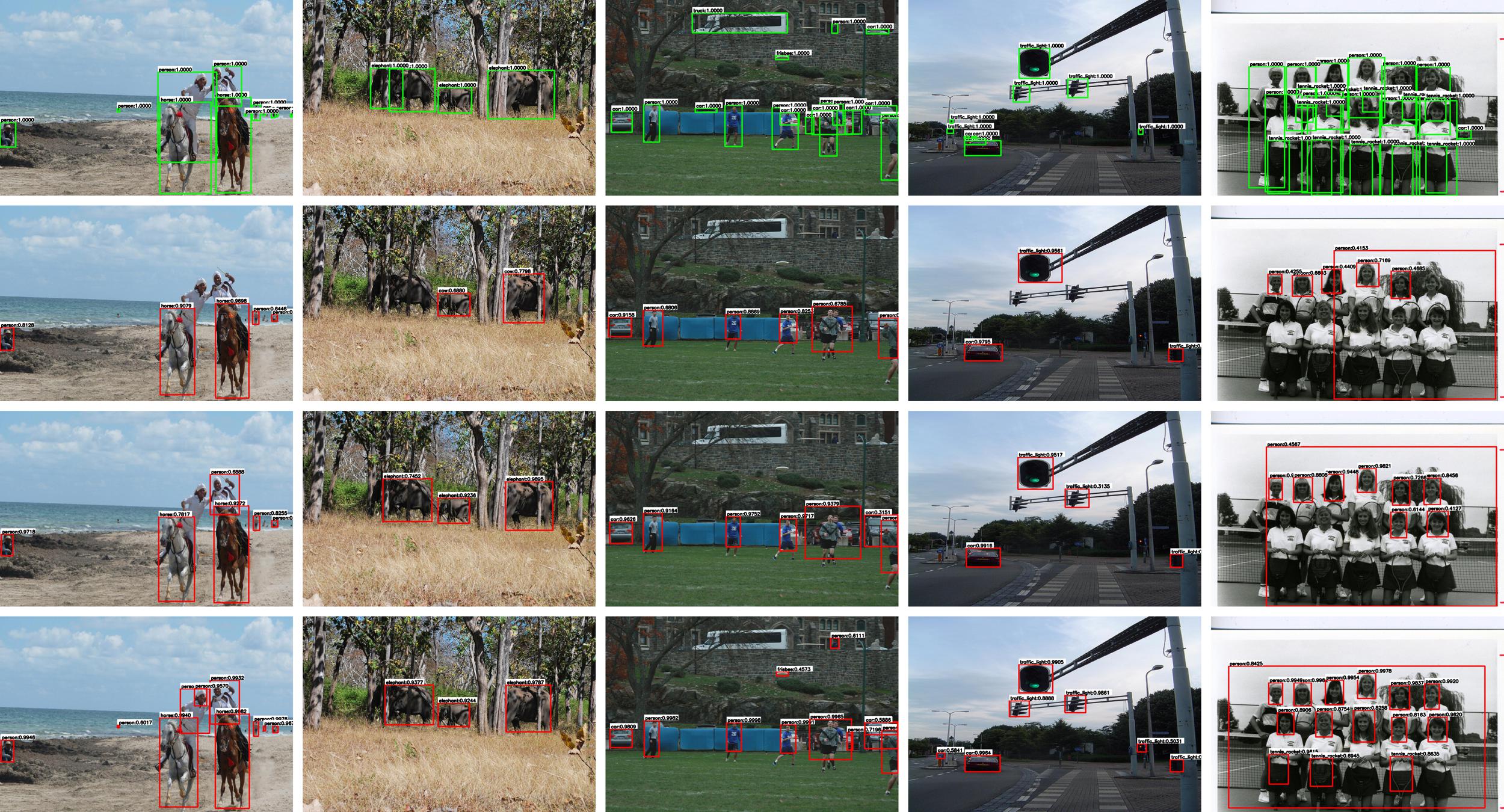}
	\caption{Visualization of SoS-WSOD results on MS-COCO. Top row: groundtruth annotations. 2nd to 4th rows: detection results from stages 1, 2 and 3, respectively. Last column: a failure case.}
	\label{fig-visualization-coco}
	\vspace{-8pt}
\end{figure*}

\begin{figure}
  \centering
  \begin{subfigure}[b]{0.49\linewidth}
    \centering
    \includegraphics[width=1\linewidth]{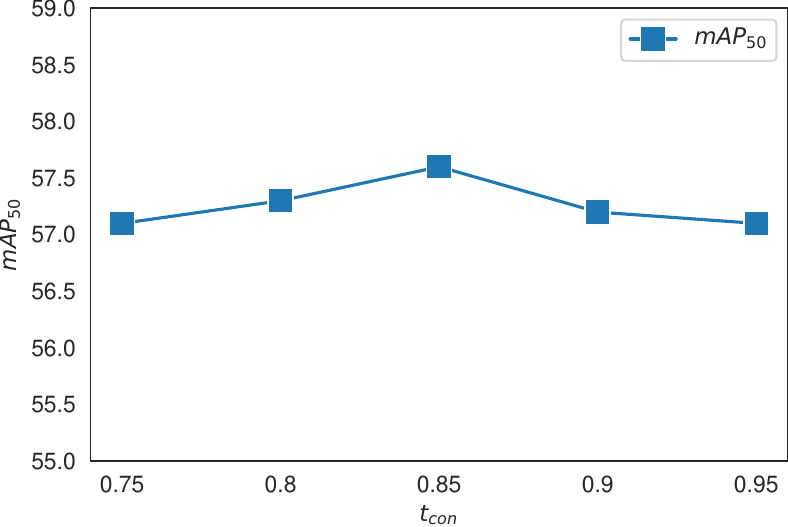}
    \caption{accuracy with various $t_{con}$.}
  \end{subfigure}
  \hfill
  \begin{subfigure}[b]{0.49\linewidth}
    \centering
    \includegraphics[width=1\linewidth]{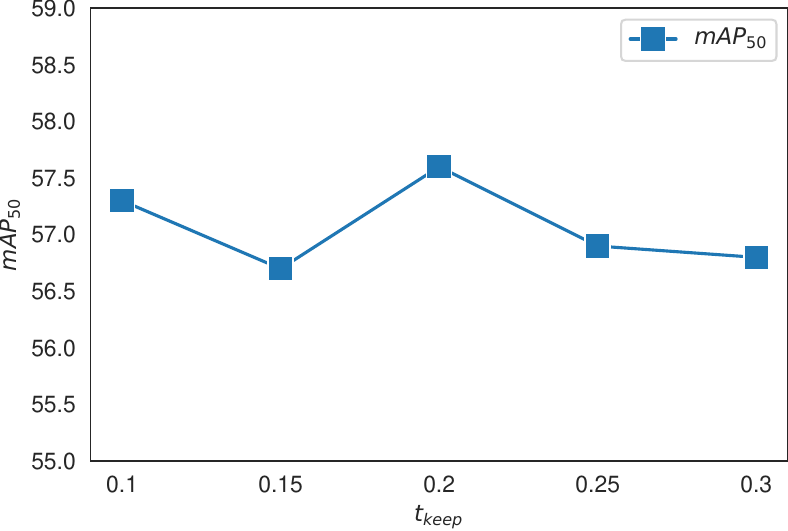}
    \caption{accuracy with various $t_{keep}$.}
  \end{subfigure}
  \caption{Effect of hyperparameters in PGF.}
  \label{fig-hyper-pgf}
    \vspace{-10pt}

\end{figure}

\textbf{Hyperparameters in PGF.} Figure~\ref{fig-hyper-pgf} shows the effects of hyperparameters $t_{keep}$ and $t_{con}$ introduced in PGF~(Algorithm \ref{alg:pgf}). These two hyperparameters are robust and $t_{keep} = 0.2, t_{con} = 0.85$ works best for $m\text{AP}_{50}$ on VOC2007. We tune these two hyperparameters on the smallest VOC2007 dataset and keep them fixed on all other datasets following previous works~\cite{casdnips2020, cmilcvpr2019, pcltpami2018}.

\textbf{Inference speed.} SoS-WSOD also enjoys speed benefits from modern FSOD methods. We compare the inference speed in Table~\ref{ablation4-speed} (on single RTX3090 GPU). Please note that the time for generating proposals is always far longer than 0.2 seconds per image, e.g., 8.3 s/img for Selective Search~\cite{selectivesearchijcv2013}, while our SoS-WSOD does not need to generate external proposals. Hence, SoS-WSOD not only is significantly faster than baseline WSOD methods, but also eliminates the time to generate external proposals.

\begin{table}
    \caption{Inference speed comparison. ``Reg'' means the bounding box regression branch.}
    \label{ablation4-speed}
    \centering
    \resizebox{0.45\textwidth}{!}{
    \begin{tabular}{l|c|c}
    \hline
    Method   & Pro.  Time~(s / img) & Inf. Time~(s / img) \\ \hline
    OICR~(+Reg.)~\cite{oicrcvpr2017} &          $> 0.2$                &    $0.101$                     \\ \hline
    SoS-WSOD & 0                        &        $0.031$                 \\ \hline
    \end{tabular}
    }
    \vspace{-10pt}
\end{table}

Finally, we provide visualization of detection results on MS-COCO in Fig.~\ref{fig-visualization-coco}. These results show that SoS-WSOD can mine more correct objects even in complicated environments. Additional visualization results on VOC2007 and MS-COCO are shown in the appendix.
\vspace{-3pt}
\section{Conclusions and Remarks}
\vspace{-1pt}

In this paper, we proposed a new three-stage framework called Salvage of Supervision for the weakly supervised object detection task~(SoS-WSOD). SoS-WSOD tackles the WSOD problem from a new perspective, which advocates harnessing all potentially useful supervisory signals (i.e., salvage of supervision) and successfully adopted modern fully supervised detection techniques in WSOD.

The first stage is a WSOD training stage, in which we train a detector with any WSOD method. Pseudo-FSOD, the second stage, improves the WSOD detector by harnessing the pseudo groundtruth generated by PGF and then freely using techniques from modern FSOD. Finally, stage 3 treats the generated pseudo-labeled dataset as a dataset with noisy labels and proposes a novel criterion to split images into labeled and unlabeled subsets, so semi-supervised detection can be used to squeeze useful supervisory signals to further improve the detection performance. Extensive experiments and visualization on VOC2007, VOC2012 and MS-COCO proved both the effectiveness of our SoS-WSOD and extra supervision signals. By successfully utilizing modern FSOD methods, SoS-WSOD can also have faster detection speed than previous WSOD methods. 

As for the limitation, SoS-WSOD still suffers from a large performance gap compared to FSOD, especially on COCO. Due to the lack of fully annotated box-level annotations, we need to salvage more supervisory signals in the future. And, SOS-WSOD still suffers problems like part domination, missing instances and clustered instances, which are widely occurred in WSOD. In the future, we will continue to explore to solve the common WSOD problem and develop better rules to split datasets and stronger SSOD methods for the WSOD task.

\clearpage

{\small
\bibliographystyle{ieee_fullname}
\bibliography{egbib}
}

\clearpage
\appendix
\setcounter{table}{0}

\renewcommand{\thesection}{A.\arabic{section}}

\section{Introducing the pipeline of OICR}

In this part, we will introduce the details of OICR~\cite{oicrcvpr2017}, a widely used framework in WSOD. OICR is composed of two parts, a multiple instance detection network~(MIDN) and several online instance classifier refinement~(OICR) branches. There are different choices to implement the MIDN part. WSDDN~\cite{wsddncvpr2016}, the first work to integrate the MIL process into an end-to-end detection model, is the most commonly used one. As for the OICR branch, originally it only contained one classifier and a softmax function. \cite{e2ewsodiccv2019} started to introduce the bounding box regressor into OICR branches, which was proved to be effective in many works~\cite{wetectroncvpr2020, casdnips2020, imcfgaaai2021, wsod2iccv2019}.

Specifically, we denote $\bm{I} \in \mathds{R}^{h \times w \times 3}$ as an RGB image, $\bm{y} = [y_1, y_2, \ldots, y_C] \in [0, 1]^{C}$ as its corresponding groundtruth class labels, and $\bm{R} \in \mathds{R}^{4 \times N}$ as the pre-computed object proposals. $C$ is the total number of object categories and $N$ is the number of proposals. With the help of a pre-trained backbone model, we can extract the feature map for $\bm{I}$, and proposal feature vectors are extracted by an RoI pooling layer and two FC layers. Following WSDDN, proposal feature vectors are  branched into two streams to produce classification logits $\bm{x}^c \in \mathds{R}^{C \times N}$ and detection logits $\bm{x}^d \in \mathds{R}^{C \times N}$. Then $\bm{x}^c$ and $\bm{x}^d$ will be normalized by passing through two softmax layers along the category direction and the proposal direction, respectively, as shown in Equation~\ref{eq: softmax}. $[\sigma(\bm{x}^{c})]_{ij}$ represents the probability of proposal $j$ belonging to class $i$ and $[\sigma(\bm{x}^{d})]_{ij}$ represents the likelihood of proposal $j$ to contain an informative part of class $i$ among all proposals in image $\bm{I}$.
\begin{equation}
    [\sigma(\bm{x}^c)]_{ij} = \frac{\exp^{x^{c}_{ij}}}{\sum_{k=1}^C \exp^{x^{c}_{kj}}}\,, \;  [\sigma(\bm{x}^d)]_{ij} = \frac{\exp^{x^{d}_{ij}}}{\sum_{k=1}^N \exp^{x^{d}_{ik}}}\,.
    \label{eq: softmax}
\end{equation}
The final proposal scores of a multiple instance detection network are computed by element-wise product: $\bm{x}^{R} = \sigma(\bm{x}^c) \odot \sigma(\bm{x}^d)$. During the training process, image score of the $c^{th}$ category $\phi_{c}$ can be obtained by summing over all proposals: $\phi_{c} = \sum_{r=1}^{N} \bm{x}_{c,r}^R$. Then the MIL classification loss is calculated by Equation~\ref{eq: milloss}.
\begin{equation}
    \mathcal{L}_{mil} = - \sum_{c=1}^{C} [y_{c} \log \phi_{c} + (1 - y_{c} \log(1 - \phi_{c}))]\,.
    \label{eq: milloss}
\end{equation}

As to the online instance classifier refinement~(OICR) branches, they are added on top of MIDN, i.e., WSDDN here. Proposal feature vectors are fed into another $K$ refinement stages and to generate classification logits $x^{k} \in \mathds{R}^{(C+1) \times N}, k \in \{1, 2, \ldots, K\}$. The $k^{th}$ branch is supervised by pseudo labels $\bm{y}^{k} \in [0, 1]^{(C+1) \times N}$, which are generated by top-score proposals of each category from the previous branch. One proposal will be encouraged to be classified as the $c$-th class only if it has high overlap with any top-score proposal of the previous OICR branch. The loss for the classifier of the $k^{th}$ branch is defined as Equation~\ref{eq: oicrloss}, where $w_{r}^{k}$ is the loss weight of proposal $r$:
\begin{equation}
    \mathcal{L}_{r}^{k} = - \frac{1}{N} \sum_{r=1}^{N} \sum_{c=1}^{C+1} w_{r}^{k} y_{c,r}^{k} \log x_{c,r}^{k}\,.
    \label{eq: oicrloss}
\end{equation}

The loss for bounding box regressor of the $k^{th}$ OICR branch is defined as Equation~\ref{eq: regressionloss}, $N_{pos}$ is the number of positive proposals in the $k^{th}$ branch, $\lambda_{reg}$ is a scalar weight of the regression loss, $t_{r}^{k}, \hat{t}_{r}^{k}$ are the predicted and pseudo groundtruth offsets of the $r^{th}$ positive proposal in the $k^{th}$ branch, respectively:
\begin{equation}
    \mathcal{L}_{reg}^{k} = \frac{1}{N_{pos}} \sum_{r=1}^{N_{pos}} \lambda_{reg} \mathcal{L}_{smooth-L1}(t_{r}^{k}, \hat{t}_{r}^{k}) \,.
    \label{eq: regressionloss}
\end{equation}

\section{Details of our improved OICR}

In this part, we provide details of our improved OICR~\cite{oicrcvpr2017}, which is used in stage 1. As we claimed in Sec.~3.1, we proposed an improved OICR as the baseline in our main experiments.

\textbf{Mining Rules.} Recent works~\cite{oimaaai2020, wetectroncvpr2020, pcltpami2018, imcfgaaai2021} demonstrate that better mining rules are critical in obtaining higher recall of objects. OICR mines proposals that have high overlap with top-scoring proposals. MIST~\cite{wetectroncvpr2020} mines more proposals with low overlap between each other but mines many wrong proposals, too. We notice that recall and precision are both essential for mining proposals. Hence, we introduce a mining rule~(Algorithm ~\ref{alg:mining step}) to strike a balance between the two factors. In Line 6, the rule to only retain the top $p$ percent of proposals is learned from MIST, but we remove low score proposals to keep the precision.

\textbf{Multi-Input.} A very recent paper CASD~\cite{casdnips2020} showed that the self-attention transfer between different versions of an input image is beneficial for boosting performance in WSOD. We find that adopting the multi-input technique alone is also helpful for performance and stability of the training process even without using inverted attention, CASD's self-attention transfer and other tricks. We randomly select inputs with two different scales and their flipped versions, feed them into the model to obtain RoI scores for different inputs, and average the scores of each proposal to get the final RoI scores.

\renewcommand{\thealgorithm}{A.\arabic{algorithm}}

\begin{algorithm}
	\caption{Mining Rules in SoS-WSOD}
	\begin{algorithmic} [1]
		\Statex {\textbf{Input}: An input image $I$, class labels $y_1, \ldots, y_m$ that are active in $I$, a set of proposals $R$ with size $n$, maximum percent $p$, score threshold $s_t$}
		\Statex {\textbf{Output}: Pseudo groundtruth seed boxes $\hat{R}$ for $I$}
		\State \quad {$\hat{R} = \varnothing$ 
		\State \quad Feed $I$ and $R$ into the model to obtain RoI scores $S$ for each proposal in $R$}
		\State \quad \textbf{for} {$i=1,\ \ldots,\ m$} \textbf{do}
		\State \quad \quad {$S_i = S[i,:]$  \quad // get scores for the $i$-th active class}
		\State \quad \quad {$R_i = \operatorname{SORTED}_{S_i}(R)$ \quad // sort the proposals according to the scores in $S_i$}
		\State \quad \quad Pick top $n \times p$ proposals, but \emph{remove those whose scores are low} ($<s_t$). Denote them as $R_i'$
		\State \quad \quad {$R'_i = \operatorname{NMS}(R'_i,0.01)$ \quad // remove those proposals having overlap with higher scored ones}
        \State \quad \quad {$\hat{R} = \hat{R}\bigcup R'_i$}
        \State \quad \textbf{end for}
	\end{algorithmic}
	\label{alg:mining step}
\end{algorithm}

\section{Implementation Details}

In this section, we provide additional implementation details for completeness.

In the WSOD training stage, we set the maximum iteration numbers to 50k, 60k and 200k for VOC2007, VOC2012 and MS-COCO, respectively. Batch size is set to be 4 for the basic OICR model. as we input 4 images with 4 different input transformations, the actual batch size is 16 when we use the improved OICR. When training the improved OICR model, $p=0.1, s_t=0.05$ are set for all datasets. When training the FSOD model with pseudo ground-truth, maximum iteration numbers are 12k, 18k, 50k for VOC2007, VOC2012 and MS-COCO, respectively. Learning rate and batch size are 0.01 and 8 for VOC2007 and VOC2012. For MS-COCO, we double the batch size to 16 and adjust the learning rate to 0.02 based on batch size. The learning rate is decayed with a factor of 10 at (8k, 10.5k), (12k, 16k) and (30k, 40k) for VOC2007, VOC2012 and MS-COCO, respectively. When mining potential useful supervisory signals by the semi-supervised learning paradigm, maximum iteration numbers are 15k, 30k, 50k for VOC2007, VOC2012 and MS-COCO, respectively. Batch sizes for the unlabeled subset and ``clean'' labeled subset are both 8 on VOC2007 and VOC2012, and doubled to 16 on MS-COCO. Learning rate is set to 0.01 on all datasets. We do not modify any other hyperparameters of object detectors.

As for the data argumentation, following~\cite{uwsod2020}, we use random flip and multi-scale training in which scales range from 480 to 1216 with stride 32 in stage 1. In stage 2 and 3, we apply the same data augmentations as~\cite{unbiasediclr2021}. For weak augmentation, only scale transform and random flip are used. Color jittering, grayscale, Gaussian blur, and cutout patches are randomly applied for strong augmentation additionally.

\section{Ability to adopt modern backbones}

In order to show that SoS-WSOD can readily enjoy the benefits from modern fully supervised object detection techniques, we conducted experiments using ResNet101 and ResNeXt101, which are widely used in fully supervised object detection, as the backbone of SoS-WSOD in stages 2 and 3. In Table~\ref{backbones-voc07}, we show the results on VOC2007. These results demonstrate that our SoS-WSOD can successfully adopt different modern backbones. Note that TTA was \emph{not} used for results in Table~\ref{backbones-voc07}.

\renewcommand{\thetable}{A.\arabic{table}}

\begin{table}[thbp]
    \centering
     \resizebox{0.45\textwidth}{!}{
    \begin{tabular}{c|c|c|c|c|c}
        \hline
        Backbone      & PGF & SSOD & $m\text{AP}_{50:95}$ & $m\text{AP}_{50}$ & $m\text{AP}_{75}$ \\ \hline
        ResNet50      &  \checkmark  &      &    27.3   &   57.6    &    22.5   \\ \hline
        ResNet50      &  \checkmark  &   \checkmark  &    31.6   &   62.7    &    28.1   \\ \hline
        ResNet101     &  \checkmark  &      &    28.7   &   58.2    &    24.2   \\ \hline
        ResNet101     &  \checkmark  &   \checkmark  &   32.4    &   63.2    &    29.3   \\ \hline
        ResNeXt101    &  \checkmark  &      &   29.1    &   59.1    &    25.5   \\ \hline
        ResNeXt101    &  \checkmark  &   \checkmark  &   33.0    &   64.7    &   30.1    \\ \hline
    \end{tabular}
    }
    \caption{Results for SoS-WSOD when using ResNet101 and ResNeXt101 as the backbone on VOC2007.}
    \label{backbones-voc07}
\end{table}

\section{Ability to adopt different detector architectures}

In order to show that SoS-WSOD can also enjoy benefits from different detector architectures, we conducted experiments using Cascade R-CNN~\cite{cascadercnn2018} with ResNet50 as the backbone on the VOC2007 dataset. Experiment results in Table~\ref{architectures-voc07} show that SoS-WSOD can successfully adopt different modern detector architectures such as Cascade R-CNN. The experimental results also illustrate that using Cascade R-CNN as the detector, SoS-WSOD can obtain performance gains and more high-quality detection results.

\begin{table}[thbp]
    \centering
     \resizebox{0.45\textwidth}{!}{
    \begin{tabular}{c|c|c|c|c|c}
        \hline
        Detector      & PGF & SSOD & $m\text{AP}_{50:95}$ & $m\text{AP}_{50}$ & $m\text{AP}_{75}$ \\ \hline
        Faster R-CNN      &  \checkmark  &      &    27.3   &   57.6    &    22.5   \\ \hline
        Faster R-CNN      &  \checkmark  &   \checkmark  &    31.6   &   62.7    &    28.1   \\ \hline
        Cascade R-CNN     &  \checkmark  &      &    29.9   &   56.7    &    27.6   \\ \hline
        Cascade R-CNN     &  \checkmark  &   \checkmark  &   32.5    &   61.3    &    30.8   \\ \hline
    \end{tabular}
    }
    \caption{Results for SoS-WSOD when using Cascade R-CNN as the detector on VOC2007.}
    \label{architectures-voc07}
\end{table}

\section{Result on VOC2012}

The results on VOC2012 we reported in Sec.~4 of the main paper were directly returned from the evaluation server of the PASCAL VOC Challenge~\cite{vocijcv2010}. The detailed results of SoS-WSOD~(using all stages) can be obtained by visiting these two anonymous result links.\footnote{\url{http://host.robots.ox.ac.uk:8080/anonymous/Q4JFTS.html}}\footnote{\url{http://host.robots.ox.ac.uk:8080/anonymous/PDK0Q9.html}}

\begin{table*}[thbp]
    \centering
    \scalebox{0.625}{
    \begin{tabular}{lcccccccccccccccccccccc}
        \hline
        \multicolumn{1}{l|}{Method}                & \multicolumn{1}{c|}{Backbone} & aero & bicy & bird & boa & bot & bus & car & cat & cha & cow & dtab & dog & hors & mbik & pers & plnt & she & sofa & trai &\multicolumn{1}{c|}{tv} & $m\text{AP}_{50}$ \\ \hline
        \multicolumn{23}{c}{Pure WSOD}                                                                                                                                                                                      \\ \hline
        \multicolumn{1}{l|}{WSDDN~\cite{wsddncvpr2016}}                 & \multicolumn{1}{c|}{VGG16}         &39.3&43.0&28.8&20.4&8.0&45.5&47.9&22.1&8.4&33.5&23.6&29.2&38.5&47.9&20.3&20.0&35.8&30.8&41.9& \multicolumn{1}{c|}{20.1}&30.2   \\
        \multicolumn{1}{l|}{OICR~\cite{oicrcvpr2017}}                  & \multicolumn{1}{c|}{VGG16}         &58.0&62.4&31.1&19.4&13.0&65.1&62.2&28.4&24.8&44.7&30.6&25.3&37.8&65.5&15.7&24.1&41.7&46.9&64.3& \multicolumn{1}{c|}{62.6}&41.2   \\
        \multicolumn{1}{l|}{PCL~\cite{pcltpami2018}}                   & \multicolumn{1}{c|}{VGG16}         &54.4&69.0&39.3&19.2&15.7&62.9&64.4&30.0&25.1&52.5&44.4&19.6&39.3&67.7&17.8&22.9&46.6&57.5&58.6& \multicolumn{1}{c|}{63.0}&43.5 \\
        \multicolumn{1}{l|}{W2F~\cite{w2fcvpr2018}}                   & \multicolumn{1}{c|}{VGG16}         &63.5&70.1&50.5&31.9&14.4&72.0&67.8&73.7&23.3&53.4&49.4&65.9&57.2&67.2&27.6&23.8&51.8&58.7&64.0& \multicolumn{1}{c|}{62.3}& 52.4 \\
        \multicolumn{1}{l|}{C-MIDN~\cite{cmidniccv2019}}                & \multicolumn{1}{c|}{VGG16}         &53.3&71.5&49.8&26.1&20.3&70.3&69.9&68.3&28.7&65.3&45.1&64.6&58.0&71.2&20.0&27.5&54.9&54.9&69.4& \multicolumn{1}{c|}{63.5}& 52.6  \\
        \multicolumn{1}{l|}{C-MIDN + FR~\cite{cmidniccv2019}}                & \multicolumn{1}{c|}{VGG16}         &54.1&74.5&56.9&26.4&22.2&68.7&68.9&74.8&25.2&64.8&46.4&70.3&66.3&67.5&21.6&24.4&53.0&59.7&68.7& \multicolumn{1}{c|}{58.9}& 53.6  \\
        \multicolumn{1}{l|}{Pred Net~\cite{prednetcvpr2019}}              & \multicolumn{1}{c|}{VGG16}         &66.7&69.5&52.8&31.4&24.7&74.5&74.1&67.3&14.6&53.0&46.1&52.9&69.9&70.8&18.5&28.4&54.6&60.7&67.1& \multicolumn{1}{c|}{60.4}& 52.9 \\
        \multicolumn{1}{l|}{SLV~\cite{slvcvpr2020}}                   & \multicolumn{1}{c|}{VGG16}         &65.6&71.4&49.0&37.1&24.6&69.6&70.3&70.6&30.8&63.1&36.0&61.4&65.3&68.4&12.4&29.9&52.4&60.0&67.6& \multicolumn{1}{c|}{64.5}& 53.5 \\
        \multicolumn{1}{l|}{SLV + FR~\cite{slvcvpr2020}}                   & \multicolumn{1}{c|}{VGG16}         &62.1&72.1&54.1&34.5&25.6&66.7&67.4&77.2&24.2&61.6&47.5&71.6&72.0&67.2&12.1&24.6&51.7&61.1&65.3& \multicolumn{1}{c|}{60.1}& 53.9 \\
        \multicolumn{1}{l|}{WSOD2~\cite{wsod2iccv2019}}                 & \multicolumn{1}{c|}{VGG16}         &65.1&64.8&57.2&39.2&24.3&69.8&66.2&61.0&29.8&64.6&42.5&60.1&71.2&70.7&21.9&28.1&58.6&59.7&52.2& \multicolumn{1}{c|}{64.8}& 53.6 \\
        \multicolumn{1}{l|}{IM-CFB~\cite{imcfgaaai2021}}                & \multicolumn{1}{c|}{VGG16}         &64.1&74.6&44.7&29.4&26.9&73.3&72.0&71.2&28.1&66.7&48.1&63.8&55.5&68.3&17.8&27.7&54.4&62.7&70.5& \multicolumn{1}{c|}{66.6}& 54.3 \\
        \multicolumn{1}{l|}{MIST~\cite{wetectroncvpr2020}}                  & \multicolumn{1}{c|}{VGG16}         &68.8&77.7&57.0&27.7&28.9&69.1&74.5&67.0&32.1&73.2&48.1&45.2&54.4&73.7&35.0&29.3&64.1&53.8&65.3& \multicolumn{1}{c|}{65.2}& 54.9 \\
        \multicolumn{1}{l|}{CASD~\cite{casdnips2020}}                  & \multicolumn{1}{c|}{VGG16}         &70.5&70.1&57.0&45.8&29.5&74.5&72.8&71.4&25.3&67.6&49.3&64.7&65.8&72.7&23.7&25.9&56.3&60.8&65.4&\multicolumn{1}{c|}{66.5}& 56.8  \\
        \multicolumn{1}{l|}{SoS-WSOD~(ours)}          & \multicolumn{1}{c|}{VGG16}      &67.4&83.1&56.2&20.2&44.6&80.9&82.0&78.7&30.3&76.0&49.5&56.6&74.9&76.1&30.1&29.7&64.1&56.6&76.7&\multicolumn{1}{c|}{72.6}&60.3 \\
        \multicolumn{1}{l|}{SoS-WSOD~(ours)}          & \multicolumn{1}{c|}{ResNet50}      &77.9&81.2&58.9&26.7&54.3&82.5&84.0&83.5&36.3&76.5&57.5&58.4&78.5&78.6&33.8&37.4&64.0&63.4&81.5&\multicolumn{1}{c|}{74.0}&64.4 \\ \hline
        \multicolumn{23}{c}{WSOD with transfer}                                                                                                                                                                             \\ \hline
        \multicolumn{1}{l|}{OCUD + FR~\cite{ocudeccv2020}}                  & \multicolumn{1}{c|}{ResNet50}      &65.5&57.7&65.1&41.3&43.0&73.6&75.7&80.4&33.4&72.2&33.8&81.3&79.6&63.0&59.4&10.9&65.1&64.2&72.7& \multicolumn{1}{c|}{67.2}& 60.2 \\ 
        
        \multicolumn{1}{l|}{LBBA~\cite{adjustericcv2021}}                  & \multicolumn{1}{c|}{VGG16}      &70.3&72.3&48.7&38.7&30.4&74.3&76.6&69.1&33.4&68.2&50.5&67.0&49.0&73.6&24.5&27.4&63.1&58.9&66.0& \multicolumn{1}{c|}{69.2}& 56.6 \\ \hline
    \end{tabular}
    }
    \caption{Per-class detection results on the VOC2007 test set.}
    \label{perclassmap}
\end{table*}

\begin{table*}[thbp]
    \centering
    \scalebox{0.625}{
    \begin{tabular}{lcccccccccccccccccccccc}
        \hline
        \multicolumn{1}{l|}{Method}                & \multicolumn{1}{c|}{Backbone} & aero & bicy & bird & boa & bot & bus & car & cat & cha & cow & dtab & dog & hors & mbik & pers & plnt & she & sofa & trai &\multicolumn{1}{c|}{tv} & $\text{CorLoc}_{50}$ \\ \hline
        \multicolumn{23}{c}{Pure WSOD}                                                                                                                                                                                      \\ \hline
        \multicolumn{1}{l|}{WSDDN~\cite{wsddncvpr2016}}                 & \multicolumn{1}{c|}{VGG16}         &65.1&58.8&58.5&33.1&39.8&68.3&60.2&59.6&34.8&64.5&30.5&43.0&56.8&82.4&25.5&41.6&61.5&55.9&65.9& \multicolumn{1}{c|}{63.7}&53.5   \\
        \multicolumn{1}{l|}{OICR~\cite{oicrcvpr2017}}                  & \multicolumn{1}{c|}{VGG16}         &81.7&80.4&48.7&49.5&32.8&81.7&85.4&40.1&40.6&79.5&35.7&33.7&60.5&88.8&21.8&57.9&76.3&59.9&75.3& \multicolumn{1}{c|}{81.4}&60.6   \\
        \multicolumn{1}{l|}{W2F~\cite{w2fcvpr2018}}                   & \multicolumn{1}{c|}{VGG16}         &85.4&87.5&62.5&54.3&35.5&85.3&86.6&82.3&39.7&82.9&49.4&76.5&74.8&90.0&46.8&53.9&84.5&68.3&79.1& \multicolumn{1}{c|}{79.9}& 70.3 \\
        \multicolumn{1}{l|}{Pred Net~\cite{prednetcvpr2019}}              & \multicolumn{1}{c|}{VGG16}         &88.6&86.3&71.8&53.4&51.2&87.6&89.0&65.3&33.2&86.6&58.8&65.9&87.7&93.3&30.9&58.9&83.4&67.8&78.7& \multicolumn{1}{c|}{80.2}& 70.9 \\
        \multicolumn{1}{l|}{SLV~\cite{slvcvpr2020}}                   & \multicolumn{1}{c|}{VGG16}         &84.6&84.3&73.3&58.5&49.2&80.2&87.0&79.4&46.8&83.6&41.8&79.3&88.8&90.4&19.5&59.7&79.4&67.7&82.9& \multicolumn{1}{c|}{83.2}& 71.0 \\
        \multicolumn{1}{l|}{SLV + FR~\cite{slvcvpr2020}}                   & \multicolumn{1}{c|}{VGG16}         &85.8&85.9&73.3&56.9&52.7&79.7&87.1&84.0&49.3&82.9&46.8&81.2&89.8&92.4&21.2&59.3&80.4&70.4&82.1& \multicolumn{1}{c|}{78.8}& 72.0 \\
        \multicolumn{1}{l|}{WSOD2~\cite{wsod2iccv2019}}                 & \multicolumn{1}{c|}{VGG16}         &87.1&80.0&74.8&60.1&36.6&79.2&83.8&70.6&43.5&88.4&46.0&74.7&87.4&90.8&44.2&52.4&81.4&61.8&67.7& \multicolumn{1}{c|}{79.9}& 69.5 \\
        \multicolumn{1}{l|}{MIST~\cite{wetectroncvpr2020}}                  & \multicolumn{1}{c|}{VGG16}         &87.5&82.4&76.0&58.0&44.7&82.2&87.5&71.2&49.1&81.5&51.7&53.3&71.4&92.8&38.2&52.8&79.4&61.0&78.3& \multicolumn{1}{c|}{76.0}& 68.8 \\

        \multicolumn{1}{l|}{SoS-WSOD~(ours)}          & \multicolumn{1}{c|}{VGG16}      &82.4&91.8&66.4&47.5&63.5&88.7&94.8&85.8&44.7&93.6&63.5&70.6&91.6&93.5&37.8&62.0&90.6&71.6&86.6&\multicolumn{1}{c|}{83.2}& 75.5\\
        \multicolumn{1}{l|}{SoS-WSOD~(ours)}          & \multicolumn{1}{c|}{ResNet50}      &89.5&93.0&71.8&49.2&72.5&88.7&93.8&88.4&54.4&94.3&70.5&70.6&93.0&95.1&39.7&70.2&89.6&74.7&88.1&\multicolumn{1}{c|}{86.3}& 78.7  \\ \hline
        \multicolumn{23}{c}{WSOD with transfer}                                                                                                                                                                             \\ \hline
        \multicolumn{1}{l|}{OCUD + FR~\cite{ocudeccv2020}}                  & \multicolumn{1}{c|}{ResNet50}      &85.8&67.5&87.1&68.6&68.3&85.8&90.4&88.7&43.5&95.2&31.6&90.9&94.2&88.8&72.4&23.8&88.7&66.1&89.7& \multicolumn{1}{c|}{76.7}& 75.2 \\ 
        
        \multicolumn{1}{l|}{LBBA~\cite{adjustericcv2021}}                  & \multicolumn{1}{c|}{VGG16}      &89.2&82.0&74.2&53.2&51.2&84.8&87.5&83.7&46.2&87.0&48.3&84.7&79.9&92.4&40.3&47.6&88.7&65.6&81.0& \multicolumn{1}{c|}{81.7}& 72.5 \\ \hline
    \end{tabular}
    }
    \caption{Correct localization (CorLoc) results on the VOC2007 trainval set.}
    \label{perclasscorloc}
\end{table*}

\renewcommand{\thefigure}{A.\arabic{figure}}

\begin{figure*}
    \centering
    \includegraphics[width=0.925\textwidth]{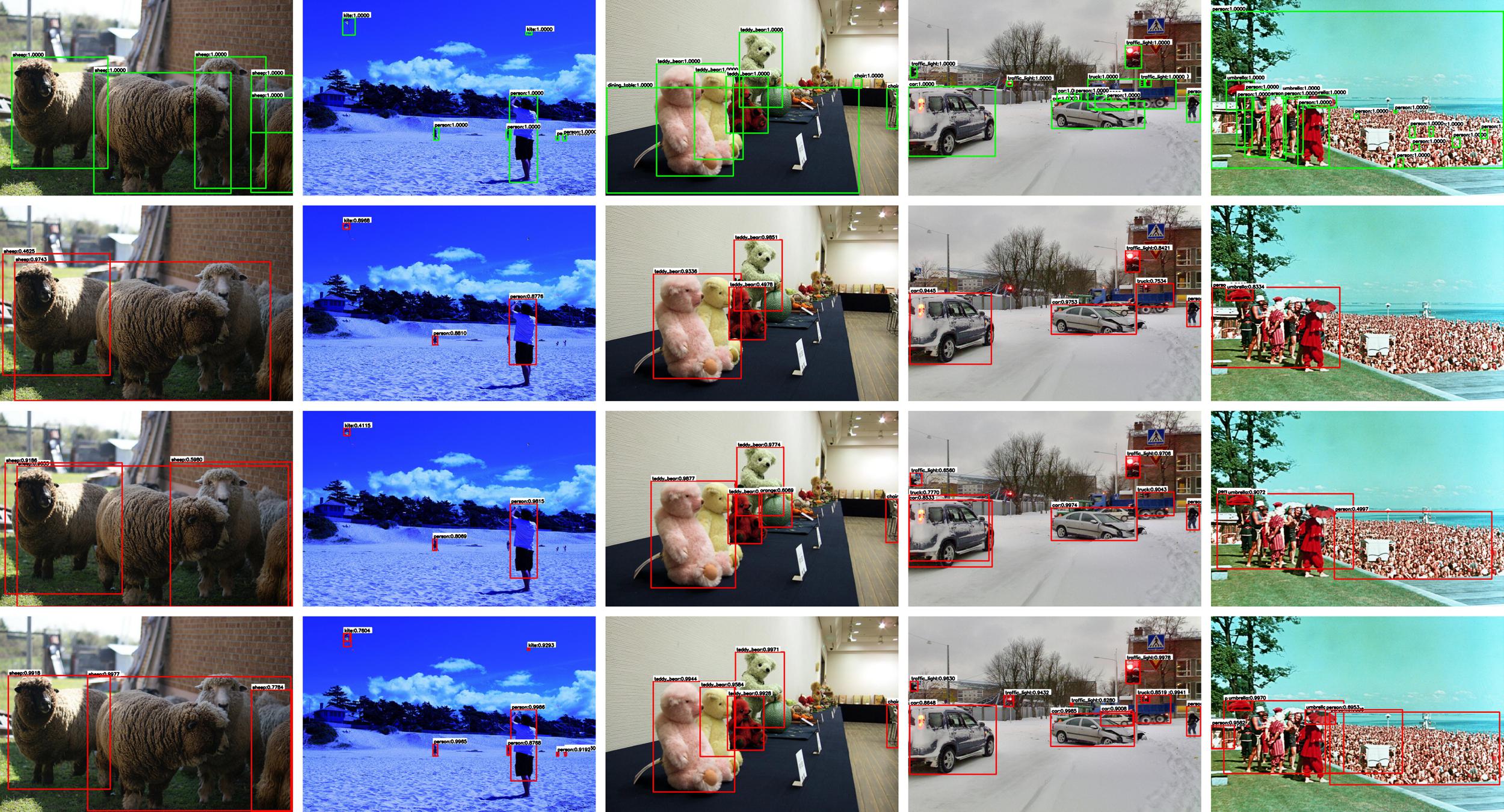}
    \caption{Visualization of SoS-WSOD results on MS-COCO (more examples in addition to Fig.~2 in the main paper). Top row: groundtruth annotations. 2nd to 4th rows: detection results from stages 1, 2 and 3, respectively. Last column: a failure case.}
    \label{fig:visualization-appendix-coco}
\end{figure*}

\begin{figure*}
    \centering
    \includegraphics[width=0.925\textwidth]{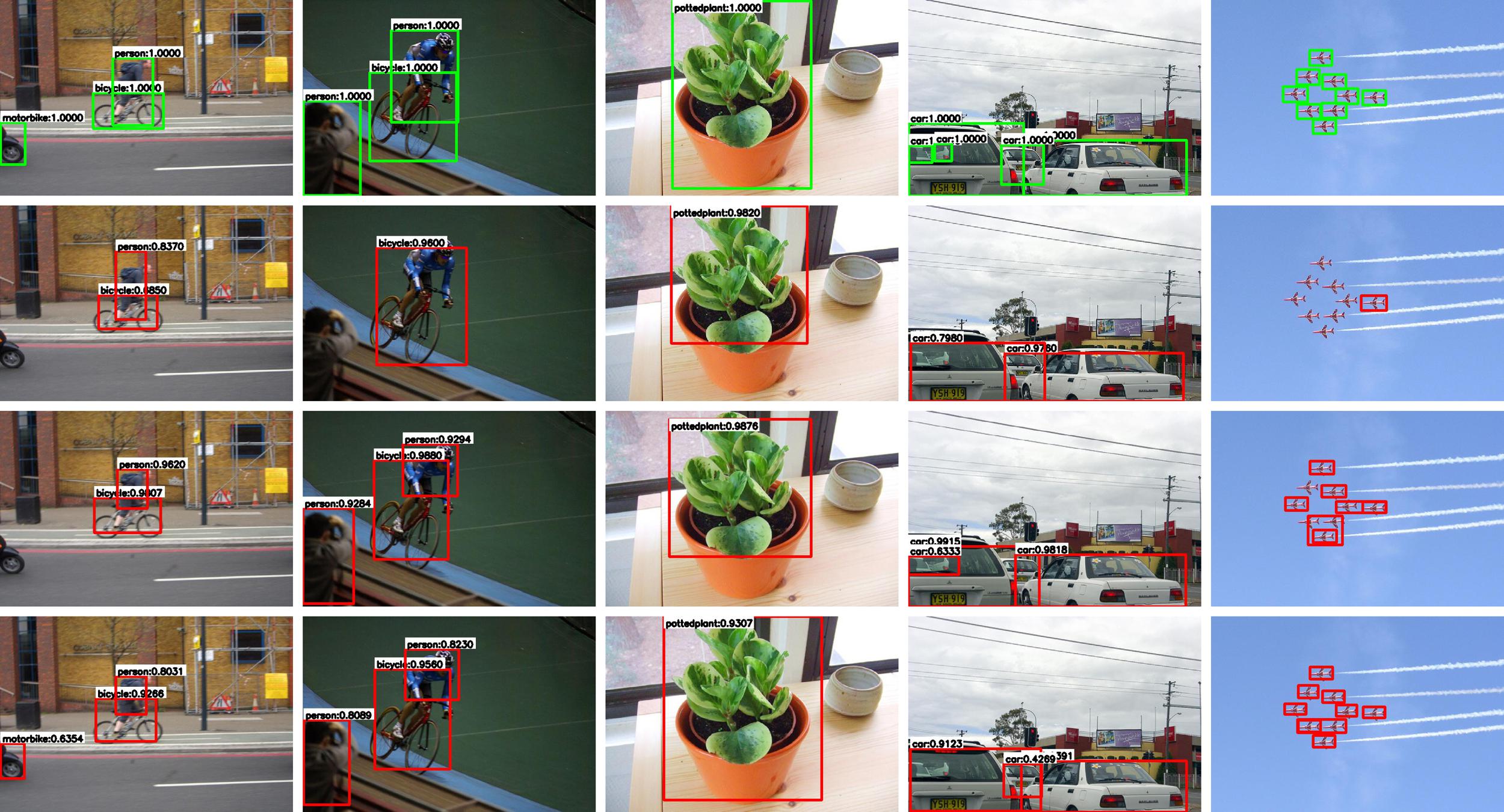}
    \caption{Visualization of SoS-WSOD results on VOC2007. Top row: groundtruth annotations. 2nd to 4th rows: detection results from stages 1, 2 and 3, respectively.}
    \label{fig:visualization-appendix-voc1}
\end{figure*}

\begin{figure*}
    \centering
    \includegraphics[width=0.925\textwidth]{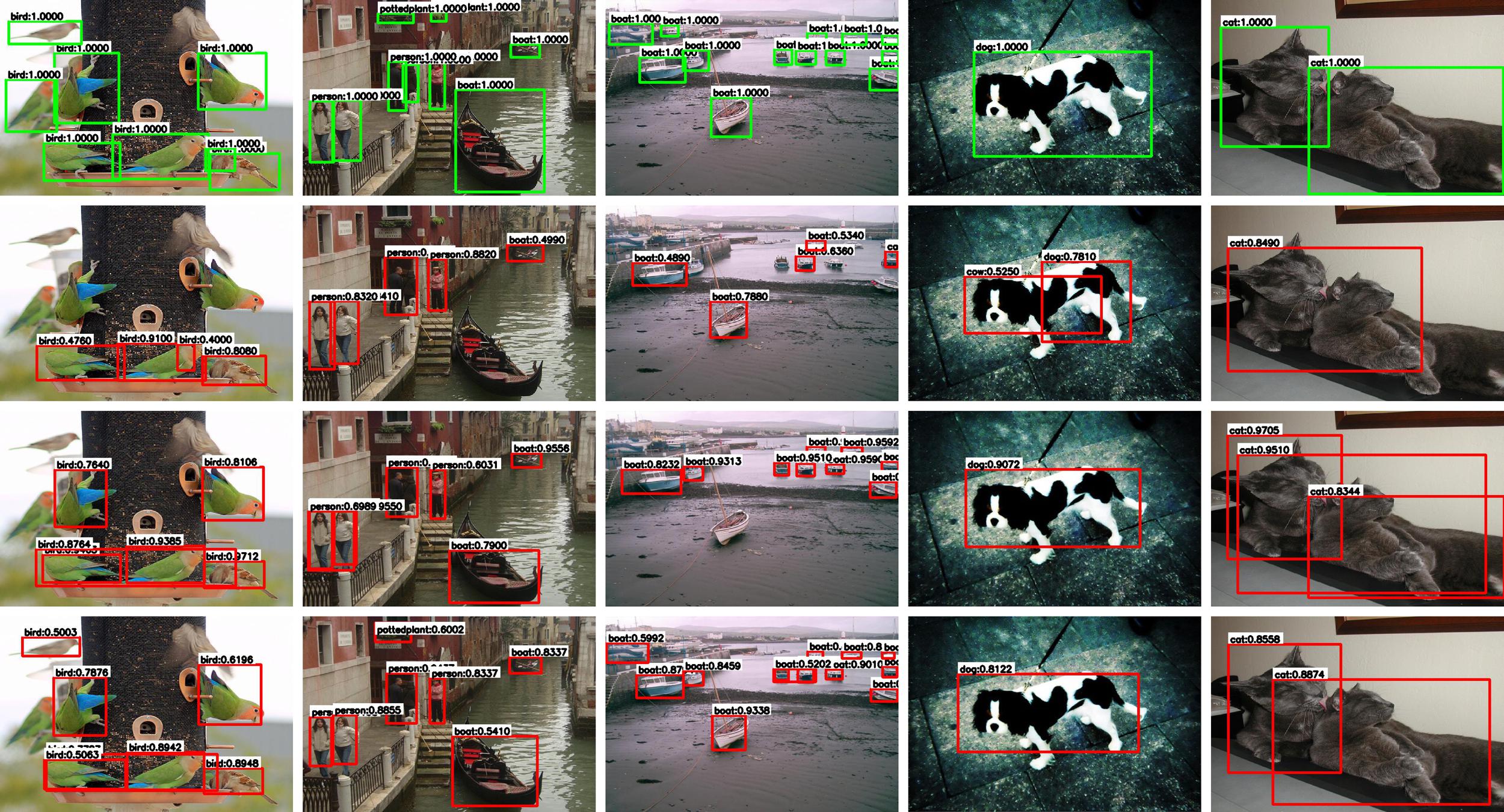}
    \caption{Visualization of SoS-WSOD results on VOC2007 (more examples in addition to Fig.~\ref{fig:visualization-appendix-voc1}). Top row: groundtruth annotations. 2nd to 4th rows: detection results from stages 1, 2 and 3, respectively.}
    \label{fig:visualization-appendix-voc2}
\end{figure*}

\clearpage

\section{Per-class detection results}

In Table~\ref{perclassmap}, we report and compare the per-class detection $m\text{AP}_{50}$ results on VOC2007. Besides, we also report and compare correct localization~(CorLoc) results on VOC2007 trainval set in Table~\ref{perclasscorloc}.

\section{More visualization results}

In Sec.~4 of the main paper, we only show some visualization results on MS-COCO due to the limited space. Here, more visualization results are shown in Fig. \ref{fig:visualization-appendix-coco} to \ref{fig:visualization-appendix-voc2}.

\end{document}